\documentclass[preprint,5pt,authoryear]{elsarticle}
\usepackage[authoryear]{natbib}
\usepackage{geometry}

\usepackage{hyperref}
\usepackage{xcolor}
\usepackage{amssymb}
\usepackage{amsmath}
\usepackage{xcolor}
\usepackage{tikz}
\usetikzlibrary{bayesnet, positioning, fit, backgrounds}
\usepackage{adjustbox}
\usepackage{colortbl}
\usepackage{xcolor}
\usepackage{multirow}
\usepackage{booktabs}

\usepackage{subcaption}
\usepackage{float}
\usepackage{booktabs}
\usepackage{amsmath}
\usepackage{amssymb}
\usepackage{array}
\usepackage{comment}

\usepackage{arydshln}
\usepackage{stfloats}

\usepackage{breqn} % rompe fórmulas automáticamente y alinea a la izquierda
\usepackage{makecell}

\usepackage{subcaption}

\usepackage{multirow}
\usepackage{bm}
\usepackage{afterpage}
\usepackage{cuted}
\usepackage{multicol}  % Para controlar las columnas manualmente
\usepackage[table]{xcolor}

\journal{Medical Image Analysis}
\makeatletter
\usepackage{xurl}
\usepackage{hyperref}

\usepackage[T1]{fontenc}
\usepackage[utf8]{inputenc}

\begin{document}
\begin{frontmatter}

\title{Systematic Morphometric Brain Autoencoding for Representation Learning in 
% Alzheimer's Disease
Neurodegenerative Diseases
\tnoteref{label1}}

\author[1]{Nuria Balbás-Esteban\corref{cor1}}

%\ead[url]{[website]}
\cortext[cor1]{Corresponding author (email: \texttt{nbalbas@ing.uc3m.es}).}
\fntext[adni]{Data used in preparation of this article were obtained from the Alzheimer's Disease Neuroimaging Initiative (ADNI) database (adni.loni.usc.edu). As such, the investigators within the ADNI contributed to the design and implementation of ADNI and/or provided data but did not participate in analysis or writing of this report. A complete listing of ADNI investigators can be found at: http://adni.loni.usc.edu/wp-content/uploads/how_to_apply/ADNI_Acknowledgement_List.pdf}
\fntext[aibl]{Data used in the preparation of this article was obtained from the Australian Imaging Biomarkers and Lifestyle flagship study of ageing (AIBL) funded by the Commonwealth Scientific and Industrial Research Organisation (CSIRO) which was made available at the ADNI database (www.loni.usc.edu/ADNI). The AIBL researchers contributed data but did not participate in analysis or writing of this report. AIBL researchers are listed at data.aibl.org.au/adni/}

\author[2,3]{Albert Belenguer-Llorens}
%\ead{[abelengu@pa.uc3m.es]}

\author[5]{Elaheh Moradi}

\author[5]{Jussi Tohka}

\author[1,4]{Vanessa Gómez-Verdejo}
% \ead{[vanessag@ing.uc3m.es]}

\author[1]{Carlos Sevilla-Salcedo}
%\ead{casevill@ing.uc3m.es}

\author[adni]{for the Alzheimer's Disease Neuroimaging Initiative}

\author[aibl]{and the Australian Imaging Biomarkers and Lifestyle flagship study of ageing}

\affiliation[1]{organization={Department of Signal Processing and Communications, Universidad Carlos III de Madrid},
            % addressline={[Address Line]}, 
            city={Leganés},
            postcode={28911}, 
            state={Madrid},
            country={Spain}}
\affiliation[2]{
    organization={Department of Health Sciences and Technology (DHEST), ETH Zurich},
    postcode={8092},
    city={Z\"urich},
    country={Switzerland}
}

\affiliation[3]{
    organization={SIB Swiss Institute of Bioinformatics},
    city={Lausanne},
    country={Switzerland}
}

\affiliation[5]{
    organization={A. I. Virtanen Institute for Molecular Sciences, University of Eastern Finland},
    city={Kuopio},
    postcode={70211},
    country={Finland}
}

\affiliation[4]{
    organization={Instituto de Investigaci\'on Sanitaria Gregorio Mara\~n\'on (IiSGM)},
    city={Madrid},
    postcode={28009},
    country={Spain}
}
% \affiliation[2]{organization={Department of [Department], [Second Institution]},
%             addressline={[Address Line]}, 
%             city={[City]},
%             postcode={[Postcode]}, 
%             state={[State]},
%             country={[Country]}}
\begin{abstract}
    Early identification of neurodegenerative disorders using structural T1-weighted MRI is essential for timely clinical intervention, yet supervised deep learning models often struggle to generalize due to limited annotations and scanner-induced distribution shift. Self-supervised learning offers a promising alternative; however, existing approaches are largely adapted from natural image paradigms, relying on contrastive objectives or generic masked reconstruction with voxel-wise losses, which may not fully capture the distributed morphometric patterns associated with neurodegeneration.
    We introduce SYMBA (SYstematic Morphometric Brain Autoencoding), a self-supervised pretraining framework specifically designed for structural brain MRI. SYMBA integrates anatomically constrained masking, a multi-scale structural similarity (MS-SSIM) reconstruction objective, and a progressive curriculum that transitions from local anatomical detail to global brain morphology. These components explicitly incorporate neuroanatomical structure into the self-supervised objective rather than relying on generic image priors.
    Pretrained on 6,953 T1-weighted MRI scans from nine heterogeneous cohorts, SYMBA is evaluated using frozen linear probing on two independent unseen datasets, enabling a strict assessment of out-of-distribution generalization. Across multiple clinical tasks, SYMBA consistently outperforms recent neuroimaging foundation models, achieving 0.930 AUROC for Alzheimer's disease classification, 0.772 for mild cognitive impairment conversion prediction, and strong performance in amyloid status prediction. In low-label regimes, SYMBA demonstrates high sample efficiency, reaching approximately 0.90 AUROC for AD/CN classification using only six labeled subjects.
    These results suggest that incorporating neuroanatomical inductive biases into self-supervised reconstruction objectives improves the transferability of learned representations and enhances label efficiency in structural MRI analysis.
\end{abstract}


\end{frontmatter}

\section{Introduction}
\label{sec:introduction}
Self-supervised learning (SSL) has emerged as a transformative paradigm in neuroimaging, enabling the extraction of robust, transferable anatomical representations directly from large collections of unlabeled brain scans \cite{li2025self, chen_foundation_2026}. A primary motivation for self-supervised representation learning in medical imaging is its ability to train effective downstream models and transfer them to very small patient populations, which is a critical bottleneck in clinical domains where expert annotations are labor-intensive, expensive, and scarce \cite{moradi2014semi}.

In this context, high-dimensional modalities like structural T1-weighted magnetic resonance imaging (MRI) provide rich volumetric depictions of brain anatomy, making them ideal for learning transferable 3D morphological representations. Neurodegeneration is an extensively studied application domain and an excellent testbed for evaluating these methodological advances. Because neurodegenerative conditions manifest as subtle, progressive structural changes distributed across interconnected brain networks \cite{jack_hypothetical_2010}, testing representation learning frameworks in this domain rigorously assesses their capacity to capture complex morphometric patterns under low-sample regimes and population heterogeneity.

%following paragraph I leave mostly unchanged:
Despite their performance on curated research cohorts \cite{wen2020convolutional}, %\cite{yang2025ensemble}
supervised DL models for neuroimaging remain difficult to translate into routine clinical practice \cite{teo2026moving}. Their reliance on large, expertly annotated datasets limits scalability, as obtaining reliable diagnostic labels is both labor-intensive and expensive. This challenge is particularly pronounced for rare or highly heterogeneous neurodegenerative disorders, where pathological variability further reduces the availability of representative training samples. In addition, supervised models are often sensitive to distribution shifts arising from differences in scanner manufacturers, acquisition protocols, and patient populations, limiting their ability to generalize across clinical sites \cite{musa2026systematic, guo_impact_2026}.

%the rest unchanged:

Recent advances in SSL, by contrast, have shown remarkable success in downstream neuroimaging tasks \cite{chen_foundation_2026} such as brain segmentation, image registration, and anatomical correspondence, where learned representations can transfer across datasets and imaging protocols. Representative examples include BrainSegFounder~\cite{cox_brainsegfounder_2024} and BrainMorph~\cite{wang_brainmorph_2025}, which leverage large-scale pretraining to improve segmentation and registration performance. However, comparatively few self-supervised pretraining frameworks have been specifically designed for neurodegenerative disease analysis, where the learning objective fundamentally differs from these structural tasks and requires capturing subtle, distributed morphometric alterations associated with disease progression rather than precise anatomical localization.

%The success of SSL has recently driven the development of foundation models (FMs) for neuroimaging. These models are pretrained on large and heterogeneous MRI repositories with the objective of learning transferable anatomical representations that remain robust across datasets and downstream tasks \cite{chen_foundation_2026}. Recent studies have demonstrated the effectiveness of this paradigm for applications such as medical image segmentation \cite{}, image registration \cite{}, and anatomical correspondence \cite{}, achieving state-of-the-art performance through large-scale pretraining. However, comparatively few studies have investigated FMs for neurodegenerative disease analysis, in which the learning objective differs substantially from that of these structural tasks \cite{ghamizi_foundation_2026}.

Although structural image segmentation and registration rely primarily on the accurate localization of anatomical boundaries, representation learning for neurodegenerative disease analysis requires capturing subtle, spatially distributed morphometric alterations that progressively affect large-scale brain networks rather than isolated brain regions \cite{akan_deep_2025, huang_magnetic_2022}. Clinically relevant biomarkers in structural MRIs, such as cortical atrophy, ventricular enlargement, or hippocampal atrophy, manifest as gradual macroscopic changes that, while initially localized, progressively extend across multiple interconnected anatomical regions as the disease evolves \cite{huang_magnetic_2022}. 
Consequently, effective self-supervised pretraining for neurodegeneration should encourage representations that model global brain morphology and long-range anatomical relationships instead of focusing solely on local structural patterns.

%Several recent FMs have begun to address this problem from different perspectives. BrainSegFounder~\cite{cox_brainsegfounder_2024} and BrainMorph~\cite{wang_brainmorph_2025} take advantage of large-scale pretraining to improve anatomical representation learning for segmentation and registration tasks, while BrainIAC~\cite{tak_foundation_2024} and AnatCL~\cite{barbano_anatomical_2025} explore contrastive learning strategies tailored to neuroimaging. These studies demonstrate the promise of FMs for brain MRI analysis, yet they also reveal that the design of effective self-supervised objectives for neurodegenerative representation learning remains far from understood.

Recent studies have begun extending SSL to neurodegenerative disease analysis. For example, BrainIAC~\cite{tak_foundation_2024} and AnatCL~\cite{barbano_anatomical_2026} demonstrate that contrastive pretraining can improve downstream diagnostic and prognostic classification tasks by learning transferable anatomical representations from large-scale MRI collections. These studies provide encouraging evidence that self-supervised representation learning is a promising direction for neurodegeneration. However, they also highlight that the design of effective pretraining objectives remains largely unexplored, with current approaches relying predominantly on contrastive learning paradigms whose relative advantages over generative objectives have yet to be systematically investigated.

%In particular, several fundamental questions remain unanswered. 
Despite these encouraging results, several fundamental questions remain unanswered.
First, although contrastive learning has become the dominant paradigm in self-supervised neuroimaging, the relative advantages of contrastive and generative pretraining objectives remain largely unexplored for neurodegenerative disease representation learning \cite{wu_towards_2025}. Second, little consensus exists regarding the most appropriate architectural choices, pretraining datasets, or evaluation protocols for assessing representation quality. Existing studies predominantly rely on end-to-end fine-tuning, making it difficult to disentangle the contribution of the learned representations from that of downstream optimization \cite{munk_towards_2026, moradi2015machine}. Finally, current generative approaches largely inherit masking strategies and reconstruction objectives originally developed for natural images. Regular cubic masking patterns and voxel-wise reconstruction losses such as Mean Squared Error (MSE) do not explicitly account for the hierarchical and spatially correlated organization of brain anatomy, potentially limiting their ability to learn morphometrically meaningful representations suitable for neurodegenerative disease analysis.

%Motivated by these challenges
We introduce SYMBA (\textbf{SY}stematic \textbf{M}orphometric \textbf{B}rain \textbf{A}utoencoding), a self-supervised pretraining framework for learning morphometric brain representations tailored to neurodegenerative disease analysis. Rather than directly adopting masked autoencoding strategies developed for natural images, SYMBA introduces an anatomically constrained masking operator that generates geometrically diverse, anatomically coherent occlusions over 3D brain volumes. The proposed masking strategy is coupled with a structural reconstruction objective based on the Multi-Scale Structural Similarity (MS-SSIM) loss \cite{wang_multiscale_2003} and a progressive masking curriculum that gradually shifts the learning task from recovering local anatomical details to modeling large-scale brain morphology. Together, these components encourage the encoder to learn biologically meaningful latent representations of brain morphology that are transferable to downstream neurodegenerative disease analysis.

%In addition, SYMBA incorporates demographic metadata to regularize the latent representation, encouraging the encoder to organize subjects according to biologically meaningful sources of variability while remaining robust to population heterogeneity. 
The resulting model is pretrained predominantly on cognitively healthy subjects from multiple large-scale cohorts, enabling learning transferable morphometric representations without relying on disease-specific supervision.

We evaluate SYMBA under a challenging transfer learning setting, where representations pretrained on healthy control populations are transferred to two independent, previously unseen neurodegenerative cohorts, including Alzheimer's disease. Performance is compared against recent self-supervised neuroimaging pretraining methods using strict frozen-representation protocols and low-label regimes, providing a controlled assessment of representation quality, transferability, and robustness under distribution shift.

\section{Methodology}
\label{sec:methodology}

\begin{figure*}[th!]
    \centering
    \includegraphics[width=1\linewidth]{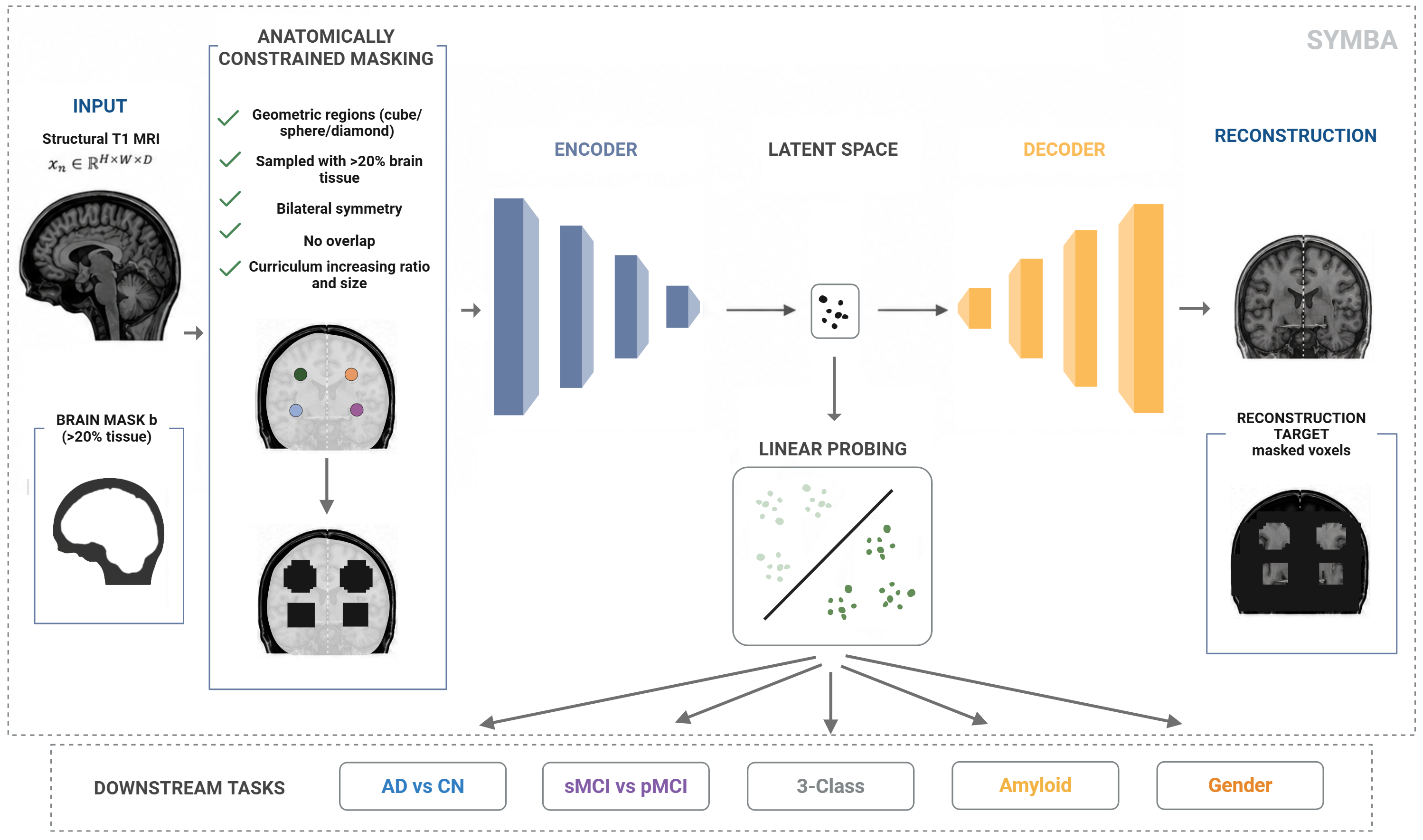} 
    \caption{Overview of SYMBA. The model learns latent representations from masked structural MRI volumes using anatomically constrained region-based masking, followed by reconstruction. The latent representations are then used to conduct linear probing on downstream tasks
    .} %and latent regularization.
    \label{fig:symba}
\end{figure*}

SYMBA is a self-supervised masked representation learning framework designed to learn anatomically meaningful latent representations from structural brain MRI without requiring diagnostic labels. Unlike conventional masked autoencoders, SYMBA adapts the masking strategy and reconstruction objective to exploit the structural properties of brain anatomy.

\subsection{Framework Overview}

Let $\mathcal{X}=\{\mathbf{x}_n\}_{n=1}^{N}$ denote a collection of unlabelled structural MRI volumes, where $N$ is the number of subjects. Each sample $\mathbf{x}_n \in \mathbb{R}^{H \times W \times D}$ represents a three-dimensional brain MRI defined over the voxel domain $\Omega=\{1,\ldots,H\}\times\{1,\ldots,W\}\times\{1,\ldots,D\}$.

For each input volume, SYMBA samples a binary masking field $\mathbf{m}_n \in \{0,1\}^{H \times W \times D}$ that removes a subset of voxels from the image, producing the masked input
\begin{equation}
\tilde{\mathbf{x}}_n=\mathbf{x}_n\odot\mathbf{m}_n,
\end{equation}
where $\odot$ denotes element-wise multiplication.

The masked image is processed by an encoder $f_{\theta}$ to obtain a compact latent representation
\begin{equation}
\mathbf{z}_n=f_{\theta}(\tilde{\mathbf{x}}_n),
\end{equation}
which is subsequently mapped by a decoder $g_{\phi}$ to reconstruct the original image
\begin{equation}
\hat{\mathbf{x}}_n=g_{\phi}(\mathbf{z}_n).
\end{equation}

Figure~\ref{fig:symba} illustrates the overall training pipeline. The masked input is first encoded into a latent representation, which is then decoded to reconstruct the original image. Training is driven by %three complementary components: anatomically constrained masking, reconstruction over the masked regions, and metadata-conditioned latent regularization. 
two complementary components: anatomically constrained masking and reconstruction over the masked regions. The following subsections describe each component in detail.

% =========================================================== %

\subsection{Anatomically Constrained Masking Operator}
\begin{figure*}[th!]
    \centering
    \includegraphics[width=\textwidth]{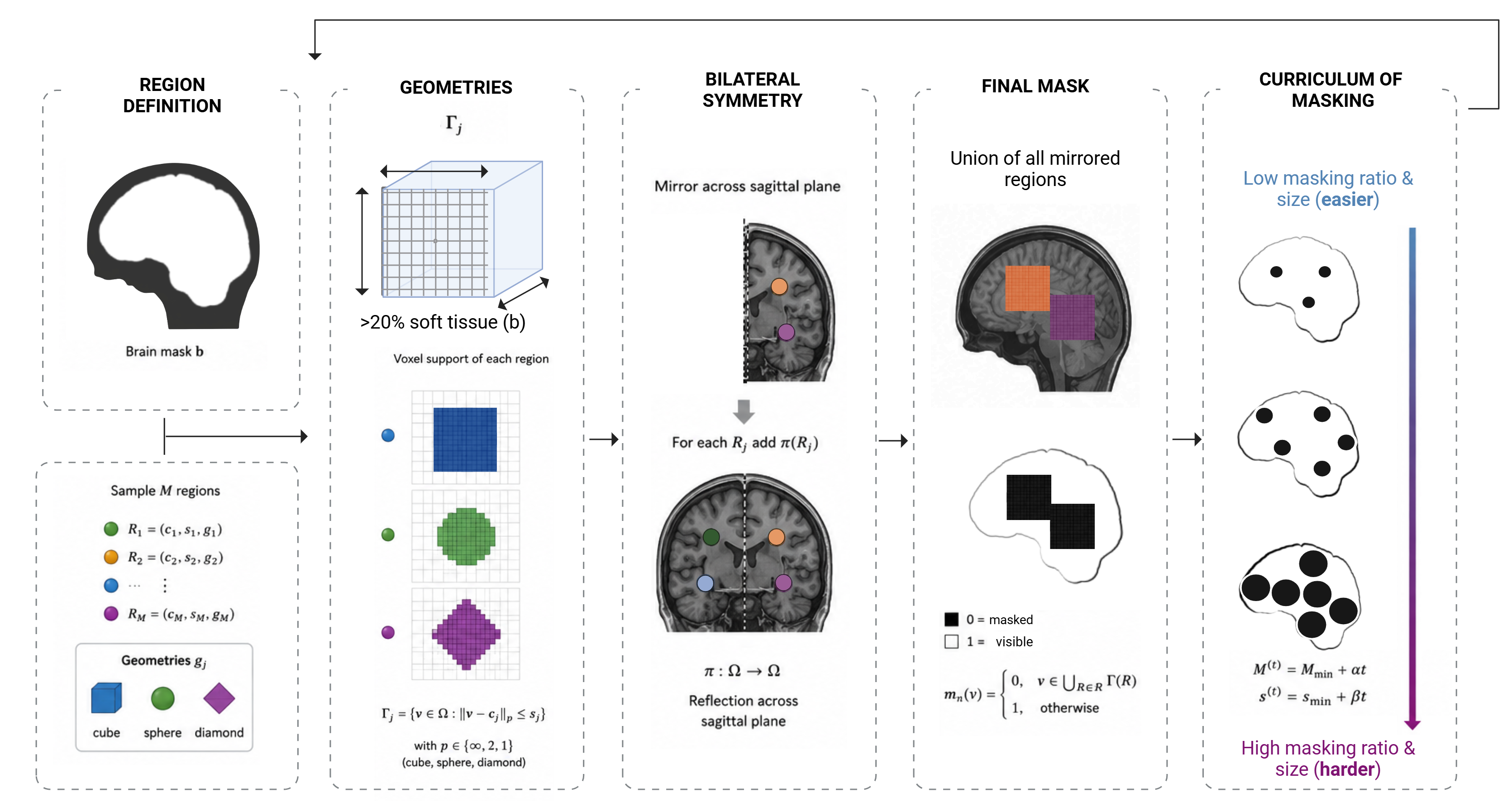}
    %\caption{Overview of the anatomically constrained masking operator. SYMBA samples geometric regions within the brain support, mirrors them across the sagittal plane, combines all region supports into a binary mask, and progressively increases masking difficulty during training.}
    \caption{Overview of the anatomically constrained masking operator used in SYMBA. The masking geometric regions are restricted to brain tissue, sampled within the brain support, and mirrored across the sagittal plane. Then, all region supports are combined into a binary mask. Masking complexity is progressively increased throughout training via curriculum learning.}
    \label{fig:masking_process}
\end{figure*}

Unlike conventional masked autoencoders, which remove arbitrary independent voxels, SYMBA constructs anatomically meaningful masking patterns by sampling geometric regions directly over the brain volume. This design encourages the model to reconstruct missing neuroanatomical structures rather than relying on fixed patch boundaries.

Given $\omega_{brain}$ as the gray matter mask, the spatial domain $\Omega$ is first partitioned into a grid of patches $\mathcal{P} = \{P_j\}_{j=1}^{K}$ of characteristic scale $s$ (representing the patch size). To ensure that masking targets meaningful neuroanatomical structures rather than empty background, patches are filtered based on their brain tissue content. We define the brain tissue support domain as
\begin{equation}
\Omega_{\mathrm{brain}} = \{\mathbf{v}\in\Omega : \mathbf{x}(\mathbf{v}) > 0\},
\end{equation}
where $\mathbf{x}(\mathbf{v})$ denotes the input volume intensity. We partition this tissue domain along the mid-sagittal plane into left and right hemispheric regions:
\begin{equation}
\Omega_{\mathrm{brain}} = \Omega_{\mathrm{brain, L}} \cup \Omega_{\mathrm{brain, R}}.
\end{equation}
For each patch $P_j$, its brain mask coverage $\tau_j^{\mathrm{L}}$ %internal brain tissue ratio 
is calculated as
\begin{equation}
\tau_j^{\mathrm{L}} = \frac{|P_j \cap \Omega_{\mathrm{brain, L}}|}{|P_j|},
\end{equation}
The masking operator restricts its candidate pool exclusively to the set of valid patches exceeding a minimum threshold:
\begin{equation}
\mathcal{P}_{\mathrm{valid, L}} = \{P_j \in \mathcal{P} : \tau_j^{\mathrm{L}} \ge 0.20\},
\end{equation}

Assuming all brain volumes are registered to a standardized stereotactic template, a symmetric masking strategy is applied to exploit the bilateral organization of brain anatomy. Every selected region is sampled on one hemisphere and then mirrored across the mid-sagittal plane. Let $\pi: \Omega_{\mathrm{brain, L}} \rightarrow \Omega_{\mathrm{brain, R}}$ denote the reflection operator across the mid-sagittal plane. We select a subset of patches $\mathcal{P}_0 \subseteq \mathcal{P}_{\mathrm{valid, L}}$ within the primary hemisphere, and construct the complete masked patch set $\mathcal{P}_{\mathrm{masked}}$ as
\begin{equation}
\mathcal{P}_{\mathrm{masked}} = \mathcal{P}_0 \cup \{\pi(P_j) : P_j \in \mathcal{P}_0\},
\end{equation}

For each batch during training, a geometric profile $g \in \{\mathrm{cube}, \mathrm{sphere}, \mathrm{diamond}\}$ is selected uniformly at random. Inside each designated patch $P_j \in \mathcal{P}_{\mathrm{masked}}$ centered at $\mathbf{c}_j$, the internal voxel support $\Gamma_j$ is determined by the profile $g$ and a radius $r = s/2$:
\begin{align}
\Gamma_j^{\mathrm{cube}}
&=
\{\mathbf{v} \in P_j : \|\mathbf{v}-\mathbf{c}_j\|_{\infty}\le r\},\\
\Gamma_j^{\mathrm{sphere}}
&=
\{\mathbf{v} \in P_j : \|\mathbf{v}-\mathbf{c}_j\|_{2}\le r\},\\
\Gamma_j^{\mathrm{diamond}}
&=
\{\mathbf{v} \in P_j : \|\mathbf{v}-\mathbf{c}_j\|_{1}\le r\}.
\end{align}

The final binary masking field $\mathbf{m}_n(\mathbf{v})$ is obtained by masking out the voxels that fall within the active geometric supports of the selected patches:
\begin{equation}
\mathbf{m}_n(\mathbf{v})
=
\begin{cases}
0,
&
\mathbf{v}\in
\displaystyle\bigcup_{P_j \in \mathcal{P}_{\mathrm{masked}}}
\Gamma_j,
\\
1,
&
\text{otherwise}.
\end{cases}
\end{equation}

Finally, masking difficulty is progressively increased during training by jointly enlarging the number of sampled regions and their characteristic scale,
\begin{equation}
M^{(t)}
=
M_{\min}
+
\alpha t,
\qquad
s^{(t)}
=
s_{\min}
+
\beta t,
\end{equation}
where $\alpha$ and $\beta$ control the curriculum schedule. This progressive masking strategy gradually transforms the learning task from local anatomical inpainting to the reconstruction of large-scale neuroanatomical structures.

Overall, the proposed masking operator combines anatomical priors, geometric diversity, bilateral symmetry, and curriculum learning to generate structured occlusions specifically tailored to structural brain MRI. The complete masking pipeline is summarized in Figure~\ref{fig:masking_process}, which illustrates the successive stages from region sampling and anatomical constraints to the construction of the final binary mask and its progressive curriculum during training.

\subsection{Reconstruction Objective}

%Given a masked input $\tilde{\mathbf{x}}_n$, the encoder--decoder network is trained to reconstruct the original MRI volume. 
Unlike conventional reconstruction objectives, which evaluate similarity over the entire image, SYMBA computes the reconstruction loss exclusively over the voxels removed by the masking operator. This prevents the model from receiving trivial supervision from already visible voxels and large homogeneous background regions, forcing it instead to recover the missing anatomical structures.

We employ the Multi-Scale Structural Similarity (MS-SSIM) index~\cite{mallick_taming_2024} to compare the reconstructed and reference masked regions. Unlike voxel-wise losses such as Mean Squared Error, MS-SSIM jointly evaluates luminance, contrast, and structural similarity across multiple spatial resolutions, making it particularly suitable for preserving anatomical morphology in structural MRI.

Given two image volumes $\mathbf{x}$ and $\hat{\mathbf{x}}$, the MS-SSIM index is defined as
%\begin{equation}
%\mathrm{MS\text{-}SSIM}(\mathbf{x},\hat{\mathbf{x}})
%=
%\left[l_M(\mathbf{x},\hat{\mathbf{x}})\right]^{\alpha_M}
%\prod_{j=1}^{L} %{M}
%\left[c_j(\mathbf{x},\hat{\mathbf{x}})\right]^{\beta_j}
%\left[s_j(\mathbf{x},\hat{\mathbf{x}})\right]^{\gamma_j},
%\end{equation}
\begin{equation}
\mathrm{MS\text{-}SSIM}(\mathbf{x},\hat{\mathbf{x}})
=
\left[l_M(\mathbf{x},\hat{\mathbf{x}})\right]^{\mu_M}
\prod_{j=1}^{L} %{M}
\left[c_j(\mathbf{x},\hat{\mathbf{x}})\right]^{\sigma_j}
\left[s_j(\mathbf{x},\hat{\mathbf{x}})\right]^{\gamma_j},
\end{equation}
where $l(\cdot,\cdot)$, $c(\cdot,\cdot)$, and $s(\cdot,\cdot)$ respectively measure luminance, contrast, and structural similarity at progressively downsampled resolutions, while $\alpha_j$, $\beta_j$, and $\gamma_j$ weight the contribution of each component.

The reconstruction objective is therefore defined as
%\begin{equation}
%\mathcal{L}_{\mathrm{recon}}
%= 1- \mathrm{MS\text{-}SSIM}
%\left(
%\mathbf{x}_n\odot(1-\mathbf{m}_n),
%\hat{\mathbf{x}}_n\odot(1-\mathbf{m}_n)
%\right).
%\end{equation}
$$\mathbf{m}_{\mathrm{loss}} = (1 - \mathbf{m}_n) \odot \mathbf{b}$$
$$\mathcal{L}_{\mathrm{recon}} = \frac{\sum_{\mathbf{v} \in \Omega} \boldsymbol{\ell}_{\mathrm{MS\text{-}SSIM}}(\mathbf{v}) \cdot \mathbf{m}_{\mathrm{loss}}(\mathbf{v})}{\sum_{\mathbf{v} \in \Omega} \mathbf{m}_{\mathrm{loss}}(\mathbf{v}) + \epsilon}$$

By restricting supervision to masked regions, the reconstruction objective complements the anatomically constrained masking strategy, encouraging the encoder to model long-range neuroanatomical dependencies rather than simply reproducing already visible tissue.

\subsection{Downstream Tasks}
% Only the encoder
% Linear probing

After self-supervised pretraining, the decoder is discarded and only the pretrained encoder $f_\theta$ is retained as a generic feature extractor. Given an input sMRI volume, the encoder produces a fixed-dimensional latent representation that serves as a compact description of the underlying brain morphology. These embeddings can subsequently be transferred to a variety of downstream tasks without requiring further representation learning.

To evaluate the quality of the learned representations, we adopt a frozen linear probing protocol. Specifically, the encoder parameters remain fixed, and a linear logistic regression classifier is trained on the extracted embeddings for each downstream task. Compared with end-to-end fine-tuning, linear probing isolates the discriminative power of the latent representation from the capacity of the downstream model, providing a more direct assessment of the information captured during self-supervised pretraining. Moreover, its simplicity reduces the influence of optimization strategies and hyperparameter choices, allowing differences in downstream performance to be more confidently attributed to the quality of the learned representation.

\section{Experiments and results}
\label{sec:results}
\subsection{Materials}
\label{sec:materials}
\begin{table*}[th!] 
\centering
\caption{Number of samples, country of origin, nature of the diagnosis and metadata of the different databases used}
\label{tab:pretraining_databases}
\resizebox{\textwidth}{!}{
\begin{tabular}{lcccccc}
\toprule
\textbf{Database} & \textbf{\# Scans } & \textbf{\# Participants} & \textbf{Country of Origin} & \textbf{Population} & \textbf{Age Ranges} \\
\midrule
Wayne & 600 & 300 & USA & Healthy & (18 - 83 years) \\
SALD & 494 & 494 & China & Healthy & (19 - 80 years) \\
IXI & 550 & 550 & UK & Healthy & (19 - 86 years)\\ 
DLBS & 315 & 315 & USA & Healthy & (21 - 89 years) \\
OASIS & 2842 & 1378 & USA & Healthy Aging \& Alzheimer's/Dementia & (42 - 95 years)\\
CamCAN & 652 & 652 & UK & Healthy & (18 - 88 years) \\
NKI & 1500 & 1500 & USA & Healthy & (6 - 85 years)\\ 
\midrule
ADNI & 3391 & 3391 & USA \& Canada & Healthy, MCI, Dementia & (50 - 91 years)\\ %- 91
AIBL & 1536 & 1536 & Australia & Healthy, MCI, Dementia & (55 - 92 years)\\ %- 90
\bottomrule
\end{tabular}
}
\end{table*}

To pretrain SYMBA, we assembled a heterogeneous structural brain MRI dataset comprising 6,953 T1-weighted scans collected from seven independent cohorts: Open Access Series of Imaging Studies (OASIS) \cite{lamontagne2019oasis}, Dallas Lifespan Brain Study (DLBS) \cite{park_dallas_2025}, IXI, SALD \cite{wei2018structural}, Wayne \cite{kennedy2009pattern}, NKI \cite{tobe2022longitudinal,musa2026systematic}, and CamCAN \cite{shafto2014cambridge}. While these cohorts each implemented highly standardized acquisition and quality-control protocols within their respective frameworks, they spanned a wide range of acquisition protocols (with varying magnetic field strengths of 1.5T and 3T), pulse sequence designs and spatial/temporal resolutions) , scanners(with manufacturers such as Siemens, GE or Philips), and subject populations. This heteroneity provided a rich %providing a diverse 
source of anatomical variability for self-supervised representation learning. A summary of the pretraining dataset is provided in Table~\ref{tab:pretraining_databases}. With the exception of OASIS, the pretraining corpus was composed predominantly of cognitively healthy individuals. 

To evaluate the transferability of the learned representations, we reserved two independent cohorts that were never used during pretraining: the Alzheimer's Disease Neuroimaging Initiative (ADNI) \cite{petersen_alzheimers_2010} and the Australian Imaging, Biomarkers and Lifestyle (AIBL) study (AIBL study methodology has been reported previously \cite{ellis2009australian}). These datasets were obtained from the ADNI (http://adni.loni.usc.edu) and the AIBL study group (https://aibl.org.au). For up-to-date information, see (www.adni-info.org). The ADNI was launched in 2003 as a public-private
partnership, led by Principal Investigator Michael W. Weiner, MD. The primary goal of ADNI has been to test whether serial magnetic resonance imaging (MRI), positron emission tomography (PET), other biological markers, and clinical and neuropsychological assessment can be combined to measure the progression of mild cognitive impairment (MCI) and early Alzheimer’s disease (AD).

These datasets were used exclusively for downstream evaluation, thereby providing a strict assessment of out-of-distribution generalization. Only baseline T1-weighted MRI scans were included in the experiments. The number of participants available for each downstream task is summarized in Table~\ref{tab:adni-performance}

%Importantly, while both independent cohorts test generalization, AIBL presented a distinct and more pronounced out-of-distribution shift relative to both the pretraining corpus and ADNI. Whereas ADNI mirrored the broad multi-site, multi-vendor hardware heterogeneity of the pretraining set, AIBL was acquired across only two primary centers with highly standardized Siemens platforms, resulting in lower inter-site variance. Furthermore, AIBL featured an older cohort ($\approx 79$ years on average) and an ethno-demographically uniform population. 
The AIBL cohort was acquired across two primary centers using standardized Siemens scanner platforms, featuring a participant group with a mean age of approximately 79 years.

To benchmark the utility of our proposed SSL approach against established SOTA SSL methods, we proposed the following tasks over the validation datasets:
\begin{enumerate}
    \item Alzheimer's Disease vs Cognitively Normal \textbf{(AD vs CN)}: Binary classification between  participants diagnosed with Alzheimer's Disease (AD) and cognitively normal controls (CN), including participants with subjective memory complaints (SMC).
    \item MCI Progressive vs Stable \textbf{(pMCI/sMCI)}: Binary classification distinguishing participants with mild cognitive impairment who subsequently progressed to Alzheimer's disease (pMCI) from those who remained clinically stable (sMCI). Stable MCI required at least five years of follow-up without progression, whereas pMCI participants were required to receive a dementia diagnosis during their final two consecutive clinical evaluations. Participants with insufficient follow-up or fluctuation in clinical assessment were excluded from the analysis.  
    \item 3-Class Categorization \textbf{(3-class)}: Three-class classification (CN, MCI, and AD) evaluating the ability of the learned representations to capture the continuum of Alzheimer's disease progression.
    \item Amyloid Status Classification \textbf{(Amyloid)}: Binary prediction of amyloid positivity ($A\beta+/-$)from structural MRI, assessing whether the learned anatomical representations encode information associated with molecular pathology. Amyloid labels were obtained using the UC Berkeley PET processing pipeline. ADNI PET acquisition and processing protocols are described in previously published methods \cite{jagust2009relationships, joshi2012performance} (\url{https://adni.loni.usc.edu/methods/pet-acquisition/} , \url{https://adni.loni.usc.edu/methods/pet-analysis-method/pet-analysis/} ). We obtained global and regional 18 F-florbetapir SUVR (standardized uptake value ratios) values from the UCBERKELEYAV45_04_26_22.csv table downloaded from the ADNI website (\url{https://adni.loni.usc.edu/} ). To determine AB-positivity, we used a cutoff value of 1.11 to the global SUVR, i.e., summary florbetapir cortical SUV normalized by whole cerebellum SUV \cite{landau2014amyloid}.
    \item Sex Classification \textbf{(Sex)}: Binary baseline mapping structural sexual dimorphism.
\end{enumerate}

Note that, due to label availability, only AD vs CN, 3-Class, and Sex include both ADNI and AIBL samples; sMCI vs pMCI and Amyloid include only ADNI samples. Table \ref{tab:adni-performance} summarizes the main information of each task.

Image preprocessing was performed using the Computational Anatomy Toolbox (CAT12, version 9.0) with default parameters \citep{gaser2024cat} \url{https://neuro-jena.github.io/cat/}. T1-weighted structural MRIs were partial volume segmented into gray matter (GM) and white matter (WM), followed by nonlinear spatial normalization to a common stereotactic space ($113 \times 137 \times 113\ 1.5\ \mathrm{mm}^3$ voxels) using the diffeomorphic shooting algorithm \citep{ashburner2011diffeomorphic}. To preserve tissue volume after spatial deformation, modulation was applied by scaling voxel intensities according to local expansion or contraction \citep{good2001voxel,kurth2015voxel}. We used stereotactically normalized gray matter density (GMD) maps as MRIs as they are comparable between participants both spatially (i.e., the voxels are spatially aligned) and intensity-wise (i.e., intensity values are GMD values that are comparable between participants \cite{kurth2015voxel}).      

\begin{table}[H]
\caption{Data split and sample distribution per task for low-data regime evaluation test.}
\centering
\renewcommand{\arraystretch}{1.2} % Proporción ideal para legibilidad de filas dobles
\setlength{\tabcolsep}{8pt}       % Espaciado óptimo entre columnas
\small                             

\begin{tabular}{llcccc}
\toprule
\multirow{ 2}{*}{\textbf{Task}} & \multirow{ 2}{*}{\textbf{Label}} & \multicolumn{2}{c}{\#\textbf{Samples}} \\ % & \textbf{Train} & \textbf{Val} & \textbf{Test} \\
\cmidrule{3-4}
& & \textbf{ADNI} & \textbf{AIBL} \\
\midrule
\multirow{2}{*}{AD vs CN}   & AD   & 395 & 69 \\ % & 197 & 99  & 99  \\
                            & CN   & 858 & 436  \\ % & 429 & 214 & 215 \\ \midrule
\multirow{2}{*}{sMCI vs pMCI}  & sMCI & 235 & -  \\ % & 117 & 59  & 59  \\
                            & pMCI & 263 & -  \\ % & 131 & 66  & 66  \\ 
\multirow{3}{*}{3-Class} & AD   & 395 & 69  \\ % & 198 & 98  & 99  \\
                            & CN   & 858 & 436  \\ % & 429 & 215 & 214 \\
                            & MCI  & 1063 & 97  \\ %& 531 & 266 & 266 \\ \midrule
\multirow{2}{*}{Amyloid}    & pos  & 581 & -  \\ % & 291 & 145 & 145 \\
                            & neg  & 539 & -  \\ % & 269 & 135 & 135 \\ \midrule
\multirow{2}{*}{Sex}     & M    & 1223 & 269  \\ %& 611 & 306 & 306 \\
                            & F    & 1104 & 336  \\ %& 552 & 276 & 276 \\ \midrule
\bottomrule
\end{tabular}
\label{tab:adni-performance}
\end{table}
Processed volumes were dynamically resized to a uniform target dimension of 128x160x128 voxels and each volume was intensity-normalized via min-max scaling to bound voxel intensities between 0 and 1. Volumes smaller than the target dimensions were symmetrically zero-padded.

\subsection{Experimental setup}

SYMBA was implemented in PyTorch. The encoder $f_\theta$ was instantiated as a 3D ResNet-50 comprising [3,4,6,3] bottleneck blocks and producing a 2048-dimensional global embedding. The decoder $g_\phi$ consisted of five transposed convolutional layers followed by a Sigmoid activation. 

SYMBA was pre-trained on the healthy-control cohorts described in Section \ref{sec:materials} together with OASIS, explicitly excluding both ADNI and AIBL from pretraining. Pretraining was performed for 600 epochs, increasing complexity through a curriculum learning technique. The masking ratio increased progressively from 30\% to 75\%, while the characteristic scale transitioned from $s^{(t)} = 16$ during the first 300 epochs to $s^{(t)} = 32$ during the remaining 300 epochs.

The network was optimized using AdamW \cite{loshchilov2019decoupledweightdecayregularization} with a weight decay of 0.05. We employed a Cosine Annealing with Warm Restarts scheduler with a peak learning rate of $1 x 10^{-4}$, with restarts synchronized with each change in curriculum complexity as done by \cite{yoon_currimae_2025}. Training used a batch size of 4, implemented via gradient accumulation. 

Reconstruction was optimized using the masked-region 3D MS-SSIM loss proposed by \cite{mallick_taming_2024}, computed over four scales. %with channel weights set to [0.0448, 0.2856, 0.3001, 0.3695]. 
Only masked brain regions satisfying $\rho\geq0.20$ were considered, and background voxels were excluded from the loss.

% For the metadata-conditioned latent regularization within the metadata-constrained model, the age RBF kernel bandwidth was set to $\sigma = 10$ years, the temperature to $\tau = 1.0$, and the scaling coefficient was evaluated at $\lambda = [0.01, 0.001]$.

We implemented a nested stratified cross-validation for each task and dataset. To provide the final performance measures we defined a 5-fold partition (outer loop), whereas for the hyperparameter tuning of the linear probe we defined a 3-folds partition (inner loop) exclusively over the training fold. 
The inner loop was used to tune the L2 regularization hyperparameter ($C \in [10^{-4}, 10^{2}]$) of a Logistic Regression classifier optimized via the L-BFGS algorithm. Performance was reported as the mean AUROC ± standard deviation of the outer loop. 

The code of both the proposed SSL approach and the experimental setup is available at \\
\href{github}{https://github.com/nuribal123/SYMBA}. However, the datasets used in this study were not hosted in the public repository due to the specific licensing constraints and terms of use associated with each database.

\subsection{Performance evaluation}
\label{subsubsec:adni-eval}

To evaluate the transferability of the learnt representations, we compared SYMBA against representative state-of-the-art large-scale, self-supervised, pre-trained models for sMRI, spanning the main SSL paradigms currently adopted in neuroimaging. Specifically, we evaluated (1) BrainMorph \cite{wang_brainmorph_2025}, a geometric representation learning approach based on anatomical keypoint prediction for image registration, (2) BrainSegFounder \cite{cox_brainsegfounder_2024}, which performed two-stage pretraining using healthy and tumor MRIs for downstream segmentation, (3) AnatCL \cite{barbano_anatomical_2026}, a contrastive framework exploiting region-of-interest information and (4) BrainIAC \cite{tak_foundation_2024}, which explored a SimCLR objective. To ensure a fair comparison, all external models were treated as frozen feature extractors and evaluated using the same linear probing protocol. Note that BrainMorph, AnatCL, and BrainIAC were originally pre-trained using ADNI as part of their pretraining datasets, whereas BrainSegFounder and SYMBA were pre-trained without any ADNI images. Consequently, while the ADNI evaluation provided a direct comparison with prior work under a common protocol, the AIBL cohort offered the strictest assessment of out-of-distribution generalisation, as none of the evaluated models were pre-trained on this dataset.

Table~\ref{tab:fm_comparison} summarises the downstream evaluation. Across both datasets, SYMBA consistently achieved the best performance on every classification task in both BACC and AUROC. Most of these improvements were statistically significant, although the margin over the strongest baselines narrowed and lost significance on the more challenging sMCI vs pMCI prediction task.

\begin{table*}[ht!]
\centering
\caption{Downstream classification performance of evaluated large-scale, self-supervised pre-trained models on the ADNI and AIBL cohorts. Results were reported as mean $\pm$ standard deviation across 10 cross-validation folds. The top row (white background) indicates Balanced Accuracy (BACC) and the bottom row (gray background) indicates AUROC (AUC). Bold indicates the best performance per task, and asterisk ($^*$) denotes statistically significant difference ($p < 0.05$) compared to SYMBA.}
\label{tab:fm_comparison}
\setlength{\tabcolsep}{2pt}
\begin{adjustbox}{max width=\textwidth}
\begin{tabular}{ccccccccc}
\toprule
& \multicolumn{5}{c}{\textbf{ADNI}} & \multicolumn{3}{c}{\textbf{AIBL}} \\
\cmidrule(lr){2-6}\cmidrule(l){7-9}
\textbf{Variant} & \textbf{AD vs CN} & \textbf{sMCI vs pMCI} & \textbf{3-Class} & \textbf{Amyloid} & \textbf{Sex} & \textbf{AD vs CN} & \textbf{3-Class} & \textbf{Sex} \\ \midrule
\multirow{2}{*}{BrainSegFounder}
			 & \cellcolor{white!100} 0.77* $\pm$ 0.04
			 & \cellcolor{white!100} 0.63* $\pm$ 0.06
			 & \cellcolor{white!100} 0.48* $\pm$ 0.03
			 & \cellcolor{white!100} 0.58* $\pm$ 0.06
			 & \cellcolor{white!100} 0.76* $\pm$ 0.04
			 & \cellcolor{white!100} 0.71 $\pm$ 0.09
			 & \cellcolor{white!100} 0.44 $\pm$ 0.07
			 & \cellcolor{white!100} 0.69* $\pm$ 0.06 \\
			 & \cellcolor{gray!10} 0.85* $\pm$ 0.03
			 & \cellcolor{gray!10} 0.68* $\pm$ 0.08
			 & \cellcolor{gray!10} 0.66* $\pm$ 0.03
			 & \cellcolor{gray!10} 0.62* $\pm$ 0.05
			 & \cellcolor{gray!10} 0.83* $\pm$ 0.04
			 & \cellcolor{gray!10} 0.81 $\pm$ 0.08
			 & \cellcolor{gray!10} 0.65 $\pm$ 0.05
			 & \cellcolor{gray!10} 0.74* $\pm$ 0.06 \\ \midrule
\multirow{2}{*}{BrainIAC}
			 & \cellcolor{white!100} 0.80* $\pm$ 0.04
			 & \cellcolor{white!100} 0.70 $\pm$ 0.06
			 & \cellcolor{white!100} 0.53* $\pm$ 0.02
			 & \cellcolor{white!100} 0.58* $\pm$ 0.04
			 & \cellcolor{white!100} 0.76* $\pm$ 0.02
			 & \cellcolor{white!100} 0.68* $\pm$ 0.06
			 & \cellcolor{white!100} 0.43 $\pm$ 0.07
			 & \cellcolor{white!100} 0.69* $\pm$ 0.05 \\
			 & \cellcolor{gray!10} 0.88* $\pm$ 0.03
			 & \cellcolor{gray!10} 0.72* $\pm$ 0.08
			 & \cellcolor{gray!10} 0.69* $\pm$ 0.02
			 & \cellcolor{gray!10} 0.63* $\pm$ 0.04
			 & \cellcolor{gray!10} 0.85* $\pm$ 0.02
			 & \cellcolor{gray!10} 0.78* $\pm$ 0.07
			 & \cellcolor{gray!10} 0.61* $\pm$ 0.05
			 & \cellcolor{gray!10} 0.76* $\pm$ 0.05 \\ \midrule
\multirow{2}{*}{BrainMorph}
			 & \cellcolor{white!100} 0.82* $\pm$ 0.03
			 & \cellcolor{white!100} 0.66* $\pm$ 0.07
			 & \cellcolor{white!100} 0.53* $\pm$ 0.04
			 & \cellcolor{white!100} 0.63 $\pm$ 0.05
			 & \cellcolor{white!100} 0.72* $\pm$ 0.04
			 & \cellcolor{white!100} 0.72 $\pm$ 0.07
			 & \cellcolor{white!100} 0.45* $\pm$ 0.06
			 & \cellcolor{white!100} 0.65* $\pm$ 0.05 \\
			 & \cellcolor{gray!10} 0.90* $\pm$ 0.03
			 & \cellcolor{gray!10} 0.74 $\pm$ 0.08
			 & \cellcolor{gray!10} 0.70* $\pm$ 0.03
			 & \cellcolor{gray!10} 0.68 $\pm$ 0.06
			 & \cellcolor{gray!10} 0.80* $\pm$ 0.03
			 & \cellcolor{gray!10} 0.86 $\pm$ 0.06
			 & \cellcolor{gray!10} 0.67 $\pm$ 0.06
			 & \cellcolor{gray!10} 0.73* $\pm$ 0.07 \\ \midrule
\multirow{2}{*}{AnatCL}
			 & \cellcolor{white!100} 0.81* $\pm$ 0.04
			 & \cellcolor{white!100} 0.68 $\pm$ 0.07
			 & \cellcolor{white!100} 0.52* $\pm$ 0.04
			 & \cellcolor{white!100} 0.62 $\pm$ 0.04
			 & \cellcolor{white!100} 0.77* $\pm$ 0.01
			 & \cellcolor{white!100} 0.70 $\pm$ 0.09
			 & \cellcolor{white!100} 0.41* $\pm$ 0.08
			 & \cellcolor{white!100} 0.67* $\pm$ 0.04 \\
			 & \cellcolor{gray!10} 0.88* $\pm$ 0.03
			 & \cellcolor{gray!10} 0.75 $\pm$ 0.07
			 & \cellcolor{gray!10} 0.68* $\pm$ 0.02
			 & \cellcolor{gray!10} 0.66 $\pm$ 0.05
			 & \cellcolor{gray!10} 0.85* $\pm$ 0.01
			 & \cellcolor{gray!10} 0.81* $\pm$ 0.07
			 & \cellcolor{gray!10} 0.65 $\pm$ 0.06
			 & \cellcolor{gray!10} 0.72* $\pm$ 0.03 \\ \midrule
\multirow{2}{*}{\textbf{SYMBA}}
			 & \cellcolor{white!100} \textbf{0.87} $\pm$ \textbf{0.04}
			 & \cellcolor{white!100} \textbf{0.71} $\pm$ \textbf{0.06}
			 & \cellcolor{white!100} \textbf{0.59} $\pm$ \textbf{0.03}
			 & \cellcolor{white!100} \textbf{0.65} $\pm$ \textbf{0.03}
			 & \cellcolor{white!100} \textbf{0.86} $\pm$ \textbf{0.02}
			 & \cellcolor{white!100} \textbf{0.78} $\pm$ \textbf{0.10}
			 & \cellcolor{white!100} \textbf{0.51} $\pm$ \textbf{0.09}
			 & \cellcolor{white!100} \textbf{0.83} $\pm$ \textbf{0.03} \\
			 & \cellcolor{gray!10} \textbf{0.94} $\pm$ \textbf{0.03}
			 & \cellcolor{gray!10} \textbf{0.78} $\pm$ \textbf{0.08}
			 & \cellcolor{gray!10} \textbf{0.74} $\pm$ \textbf{0.02}
			 & \cellcolor{gray!10} \textbf{0.70} $\pm$ \textbf{0.04}
			 & \cellcolor{gray!10} \textbf{0.94} $\pm$ \textbf{0.01}
			 & \cellcolor{gray!10} \textbf{0.88} $\pm$ \textbf{0.09}
			 & \cellcolor{gray!10} \textbf{0.71} $\pm$ \textbf{0.08}
			 & \cellcolor{gray!10} \textbf{0.92} $\pm$ \textbf{0.03} \\
\bottomrule
\end{tabular}
\end{adjustbox}
\end{table*}

On ADNI, SYMBA improved AD vs CN discrimination over the best baseline (0.82 to 0.87 BACC, 0.90 to 0.94 AUROC) while simultaneously obtaining the highest performance on the sMCI vs pMCI prediction task (0.71 BACC, 0.78 AUROC). The improvements extended to 3-Class diagnosis and amyloid prediction.

The performance gains were maintained in the AIBL cohort, where SYMBA again achieved the highest BACC and AUROC across all tasks, with significant differences over most baselines. 

\subsection{Low-data regime evaluation}
\label{subsubsec:scarcity}

To evaluate the sample efficiency of the learned representations, we progressively reduced the amount of labelled training data available for downstream learning. For each task, the training set was randomly subsampled to 1$\%$, 2$\%$, 8$\%$, 25$\%$, 50$\%$, 75$\%$, and 100$\%$ of the available participants, while the test partition remained fixed. Each subsampling was repeated 10 times per fold to account for both data subsampling and optimization variability.

We compared SYMBA against two categories of baselines. The first was the same backbone trained end-to-end under full supervision from random initialization (no pretraining), following \cite{tak_generalizable_2026}. The second was the set of four publicly available pre-trained neuroimaging models (AnatCL, BrainIAC, BrainMorph, and BrainSegFounder) evaluated under the same frozen-encoder, linear-probe protocol as SYMBA. This second comparison isolated whether SYMBA's advantage came from pretraining in general or from the specific self-supervised objective proposed here.

\begin{figure}[H]
    \vspace{-2cm}
    \centering
    \includegraphics[width=0.74\linewidth]{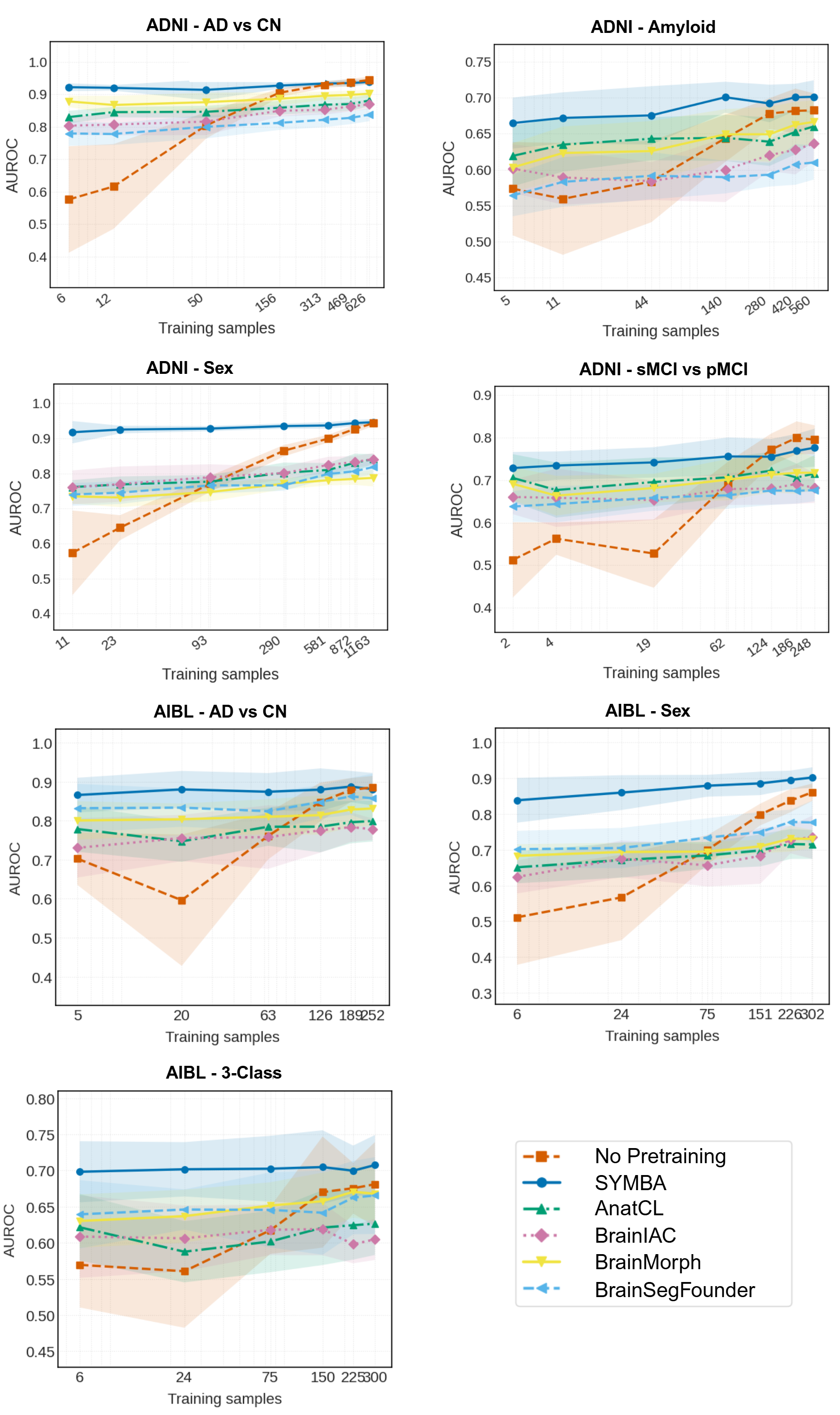} 
    %Figures/brains_spheres.png
    \caption{Downstream classification performance (AUROC) as a function of the proportion of labelled training data available, shown on a logarithmic x-axis. (a) Classification tasks A) AD vs\ CN, B) sMCI vs pMCI, C) Amyloid, D) Sex. (b)  for the AD vs\ CN and Sex tasks. In both panels, SYMBA corresponds to a frozen encoder pre-trained using the proposed self-supervised framework with a linear classifier; No Pretraining denotes the same architecture trained from random initialization under full supervision; and AnatCL, BrainIAC, BrainMorph, and BrainSegFounder are publicly available pre-trained neuroimaging foundation models evaluated under the same frozen-encoder protocol as SYMBA.}
    \label{fig:scarcity}
\end{figure}

Figure~\ref{fig:scarcity}(a) shows that on AD vs CN, Sex, and Amyloid classification, SYMBA consistently outperformed both the supervised baseline and all four alternative large pre-trained models on the ADNI dataset, with the advantage being most notable when only a handful of labelled participants were available. Strikingly, with only five to six labelled samples, SYMBA already outperformed all four large pre-trained models on AD vs CN classification, despite three of them having encountered ADNI itself during their own pretraining. As more labels were added, the supervised baseline catched up, and the two methods converged by the time the full training set was used, whereas the alternative pre-trained models plateaued at a lower level throughout, never closing the gap to either SYMBA or the fully trained supervised model.

The sMCI vs pMCI task departed from this pattern. SYMBA still provided the clearest advantage in the extreme low-data regime and remained ahead of all four alternative foundation models throughout. The supervised baseline, however, did not merely close the gap with increasing data: it overtook SYMBA once a moderate amount of labelled data became available and stayed ahead through the full training set. 

To assess whether these findings generalised beyond the pretraining cohort, we repeated the evaluation on the independent AIBL dataset for the three tasks available in it, AD vs CN, 3-Class, and sex classification (Figure~\ref{fig:scarcity}(b)). SYMBA remained the top-performing method across the full range of labelled data, and, unlike on ADNI, the fully supervised baseline did not fully catch up even at 100\% of the data. The four alternative large pre-trained models again underperformed both SYMBA and the supervised baseline at essentially every data fraction. This was particularly notable for BrainMorph, which had already seen AIBL during its own pretraining and yet still trailed SYMBA on this cohort. The cross-cohort replication indicated that SYMBA's representations transferred to genuinely unseen data rather than relying on prior exposure to the evaluation cohort.

\subsection{Ablation study}

To quantify the contribution of each design choice in SYMBA, we performed a controlled ablation study along three key axes of the proposed framework: (i) the autoencoder architecture, (ii) the pre-training paradigm, and (iii) the reconstruction objective. In each experiment, we isolated one design axis while keeping the remaining training configuration fixed, enabling us to assess its individual impact on downstream performance. This setup provided a systematic evaluation of how architectural and training choices affect the quality and transferability of the learned representations.

\subsubsection{Autoencoder architecture}

We first analyzed the impact of the autoencoder architecture on representation quality by comparing three 3D architectures that differed in both model capacity and connectivity patterns: a large 3D U-Net with encoder-decoder skip connections (UNet L, ~142.1M parameters), a medium-sized residual autoencoder (ResNet M, ~63.4M parameters), and a larger residual autoencoder with increased representational capacity (ResNet L, ~116.8M parameters). While ResNet M used a 3D-ResNet34 architecture with BasicBlock3D layers, ResNet L scaled up to a RD-ResNet50 backbone featuring Bottleneck3D layers. This comparison allowed us to disentangle the effect of network size from that of skip connections, which may facilitate low-level information transfer between encoder and decoder but can also reduce the pressure on the latent representation to capture semantically meaningful features.

%THIS IS THE GOOD ONE:
\begin{table*}[ht!]
\centering
\caption{Comparison of autoencoder architectures evaluated on ADNI and AIBL cohorts at Epoch 200. Performance was evaluated via linear probing across 10-fold cross-validation and reported as mean $\pm$ standard deviation.
For each architecture, the top row (white background) reports Balanced Accuracy (BACC) and the bottom row (shaded gray) reports AUROC.
\textbf{Bold} indicates the best performance per task and metric. Asterisks ($*$) denote statistically significant differences ($p < 0.05$, paired $t$-test) compared to the ResNet L configuration.}
\label{tab:architectures}
\setlength{\tabcolsep}{2pt}
\begin{adjustbox}{max width=\textwidth}
\begin{tabular}{ccccccccc}
\toprule
& \multicolumn{5}{c}{\textbf{ADNI}} & \multicolumn{3}{c}{\textbf{AIBL}} \\
\cmidrule(lr){2-6}\cmidrule(l){7-9}
\textbf{Architecture} & \textbf{AD vs CN} & \textbf{sMCI vs pMCI} & \textbf{3-Class} & \textbf{Amyloid} & \textbf{Sex} & \textbf{AD vs CN} & \textbf{3-Class} & \textbf{Sex} \\ \midrule
\multirow{2}{*}{UNet L} 
			 & \cellcolor{white!100} 0.83* $\pm$ 0.04
			 & \cellcolor{white!100} 0.66* $\pm$ 0.08
			 & \cellcolor{white!100} 0.53* $\pm$ 0.04
			 & \cellcolor{white!100} 0.65 $\pm$ 0.06
			 & \cellcolor{white!100} 0.85* $\pm$ 0.02
			 & \cellcolor{white!100} 0.79 $\pm$ 0.08
			 & \cellcolor{white!100} 0.53 $\pm$ 0.07
			 & \cellcolor{white!100} 0.81 $\pm$ 0.04 \\

			 & \cellcolor{gray!10} 0.91* $\pm$ 0.03
			 & \cellcolor{gray!10} 0.73* $\pm$ 0.06
			 & \cellcolor{gray!10} 0.70* $\pm$ 0.01
			 & \cellcolor{gray!10} 0.70 $\pm$ 0.06
			 & \cellcolor{gray!10} 0.92* $\pm$ 0.01
			 & \cellcolor{gray!10} \textbf{0.91} $\pm$ \textbf{0.06}
			 & \cellcolor{gray!10} 0.70 $\pm$ 0.05
			 & \cellcolor{gray!10} 0.90 $\pm$ 0.02 \\
             
\multirow{2}{*}{ResNet M} 
			 & \cellcolor{white!100} 0.85* $\pm$ 0.04
			 & \cellcolor{white!100} 0.71 $\pm$ 0.07
			 & \cellcolor{white!100} 0.56* $\pm$ 0.03
			 & \cellcolor{white!100} 0.65 $\pm$ 0.03
			 & \cellcolor{white!100} 0.81* $\pm$ 0.02
			 & \cellcolor{white!100} \textbf{0.80} $\pm$ \textbf{0.09}
			 & \cellcolor{white!100} \textbf{0.57} $\pm$ \textbf{0.07}
			 & \cellcolor{white!100} 0.82 $\pm$ 0.06 \\

			 & \cellcolor{gray!10} 0.91* $\pm$ 0.04
			 & \cellcolor{gray!10} 0.77 $\pm$ 0.06
			 & \cellcolor{gray!10} 0.72* $\pm$ 0.02
			 & \cellcolor{gray!10} 0.68 $\pm$ 0.02
			 & \cellcolor{gray!10} 0.90* $\pm$ 0.02
			 & \cellcolor{gray!10} 0.90 $\pm$ 0.06
			 & \cellcolor{gray!10} 0.71 $\pm$ 0.06
			 & \cellcolor{gray!10} 0.88* $\pm$ 0.03 \\

\multirow{2}{*}{ResNet L} 
			 & \cellcolor{white!100} \textbf{0.87} $\pm$ \textbf{0.04}
			 & \cellcolor{white!100} \textbf{0.71} $\pm$ \textbf{0.06}
			 & \cellcolor{white!100} \textbf{0.59} $\pm$ \textbf{0.03}
			 & \cellcolor{white!100} \textbf{0.65} $\pm$ \textbf{0.03}
			 & \cellcolor{white!100} \textbf{0.86} $\pm$ \textbf{0.02}
			 & \cellcolor{white!100} 0.78 $\pm$ 0.10
			 & \cellcolor{white!100} 0.51 $\pm$ 0.09
			 & \cellcolor{white!100} \textbf{0.83} $\pm$ \textbf{0.03} \\

			 & \cellcolor{gray!10} \textbf{0.94} $\pm$ \textbf{0.03}
			 & \cellcolor{gray!10} \textbf{0.78} $\pm$ \textbf{0.08}
			 & \cellcolor{gray!10} \textbf{0.74} $\pm$ \textbf{0.02}
			 & \cellcolor{gray!10} \textbf{0.70} $\pm$ \textbf{0.04}
			 & \cellcolor{gray!10} \textbf{0.94} $\pm$ \textbf{0.01}
			 & \cellcolor{gray!10} 0.88 $\pm$ 0.09
			 & \cellcolor{gray!10} \textbf{0.71} $\pm$ \textbf{0.08}
			 & \cellcolor{gray!10} \textbf{0.92} $\pm$ \textbf{0.03} \\

\bottomrule
\end{tabular}
\end{adjustbox}
\end{table*}

Table~\ref{tab:architectures} shows that, within the residual family, increasing capacity from ResNet M to ResNet L consistently improves downstream performance on ADNI, with ResNet L achieving the best results all five tasks. UNet L, despite having more parameters than either residual model, did not follow this trend: it was the weakest architecture on most ADNI tasks, with the exception of Amyloid and sex classification, where it ranked second to ResNet L. 

%In the AIBL cohort, this pattern shifted: UNet L was the strongest architecture for Sex classification (both BACC and AUROC) and 3-Class AUROC, and it was competitive with ResNet L on AD vs CN, narrowly trailing on BACC but leading on AUROC. 
In the AIBL cohort, this pattern shifted: UNet L was the strongest architecture for AD vs CN (BACC), and it was competitive with ResNet L on 3-Class, narrowly trailing on BACC but leading on AUROC.

\subsubsection{Masking strategy and SSL paradigm}

We decoupled the analysis of SYMBA into two: (i) the choice of self-supervised pre-training paradigm, and (ii) the effect of the proposed anatomically constrained masking design. This separation allowed us to isolate whether performance gains arise from the general learning objective or from the masking strategy.

We first compared contrastive learning (SimCLR, using either a standard ResNet3D backbone or a BrainIAC-style encoder) against reconstruction-based pretraining under the proposed curriculum schedule, evaluated both with (SYMBA) and without (MAE) the anatomical shape-prior term. The latter served as the reconstruction-only counterpart to the full SYMBA objective, isolating the specific contribution of the shape prior within an otherwise identical curriculum-based training scheme.

To ensure a strictly controlled comparison across self-supervised paradigms, all objectives in Table~\ref{tab:ssl_paradigms} used a smaller backbone architecture (ResNet-Small). This modification was required by GPU VRAM limitations when managing contrastive methods like SimCLR that required large batch sizes. While SYMBA employed a larger backbone (ResNet-Large) in primary experiments throughout this paper, this uniform ResNet-Small setup ensured that observed performance differenced stem entirely from the pretraining objectives rather than architectural capacity.

As shown in Table~\ref{tab:ssl_paradigms}, the full objective with the anatomical shape prior achieved the best or joint-best performance across most tasks and both cohorts, with its advantage over the contrastive baselines most pronounced on sex classification. Reconstruction-based pretraining without the shape prior (MAE) improved modestly over the contrastive baselines but fell consistently short of the full SYMBA objective.

The exception to this pattern was amyloid status and 3-class discrimination on the external AIBL cohort, where plain SimCLR remained competitive with, or marginally ahead of, the shape-prior objective. 

\begin{table*}[ht!]
\centering
\caption{Performance comparison of self-supervised representations across downstream tasks on ADNI and AIBL cohorts. Results were evaluated via linear probing and reported as mean $\pm$ standard deviation across 10-fold cross-validation. For each model configuration, the top row (white background) reports Balanced Accuracy (BACC), while the bottom row (shaded gray) reports Area Under the ROC Curve (AUC). The best performance for each task and metric is in \textbf{bold}. Asterisks ($*$) denote statistically significant differences ($p < 0.05$, paired $t$-test) compared to the baseline (SYMBA).
}
\label{tab:ssl_paradigms}
\setlength{\tabcolsep}{2pt}
\begin{adjustbox}{max width=\textwidth}
\begin{tabular}{ccccccccc}
\toprule
& \multicolumn{5}{c}{\textbf{ADNI}} & \multicolumn{3}{c}{\textbf{AIBL}} \\
\cmidrule(lr){2-6}\cmidrule(l){7-9}
\textbf{Pretraining Objective} & \textbf{AD vs CN} & \textbf{sMCI vs pMCI} & \textbf{3-Class} & \textbf{Amyloid} & \textbf{Sex} & \textbf{AD vs CN} & \textbf{3-Class} & \textbf{Sex} \\ \midrule
\multirow{2}{*}{SimCLR} 
			 & \cellcolor{white!100} 0.79* $\pm$ 0.04
			 & \cellcolor{white!100} 0.66* $\pm$ 0.06
			 & \cellcolor{white!100} 0.55 $\pm$ 0.03
			 & \cellcolor{white!100} \textbf{0.63} $\pm$ \textbf{0.04}
			 & \cellcolor{white!100} 0.69* $\pm$ 0.03
			 & \cellcolor{white!100} \textbf{0.73} $\pm$ \textbf{0.09}
			 & \cellcolor{white!100} \textbf{0.49} $\pm$\textbf{ 0.07}
			 & \cellcolor{white!100} 0.65* $\pm$ 0.06 \\

			 & \cellcolor{gray!10} 0.87* $\pm$ 0.04
			 & \cellcolor{gray!10} 0.74 $\pm$ 0.07
			 & \cellcolor{gray!10} 0.71 $\pm$ 0.02
			 & \cellcolor{gray!10} \textbf{0.68} $\pm$ \textbf{0.03}
			 & \cellcolor{gray!10} 0.76* $\pm$ 0.02
			 & \cellcolor{gray!10} 0.82* $\pm$ 0.08
			 & \cellcolor{gray!10} \textbf{0.69} $\pm$ \textbf{0.06}
			 & \cellcolor{gray!10} 0.71* $\pm$ 0.07 \\

\multirow{2}{*}{BrainIAC} 
			 & \cellcolor{white!100} 0.76* $\pm$ 0.05
			 & \cellcolor{white!100} 0.65* $\pm$ 0.08
			 & \cellcolor{white!100} 0.50* $\pm$ 0.04
			 & \cellcolor{white!100} 0.59* $\pm$ 0.03
			 & \cellcolor{white!100} 0.64* $\pm$ 0.02
			 & \cellcolor{white!100} 0.68* $\pm$ 0.09
			 & \cellcolor{white!100} 0.47 $\pm$ 0.07
			 & \cellcolor{white!100} 0.65* $\pm$ 0.08 \\

			 & \cellcolor{gray!10} 0.83* $\pm$ 0.04
			 & \cellcolor{gray!10} 0.70* $\pm$ 0.07
			 & \cellcolor{gray!10} 0.66* $\pm$ 0.03
			 & \cellcolor{gray!10} 0.62* $\pm$ 0.03
			 & \cellcolor{gray!10} 0.70* $\pm$ 0.02
			 & \cellcolor{gray!10} 0.79* $\pm$ 0.07
			 & \cellcolor{gray!10} 0.65* $\pm$ 0.05
			 & \cellcolor{gray!10} 0.69* $\pm$ 0.06 \\

\multirow{2}{*}{MAE} 
			 & \cellcolor{white!100} 0.80* $\pm$ 0.06
			 & \cellcolor{white!100} 0.66 $\pm$ 0.07
			 & \cellcolor{white!100} 0.52* $\pm$ 0.05
			 & \cellcolor{white!100} 0.62 $\pm$ 0.03
			 & \cellcolor{white!100} 0.77* $\pm$ 0.03
			 & \cellcolor{white!100} \textbf{0.73} $\pm$ \textbf{0.10}
			 & \cellcolor{white!100} 0.47 $\pm$ 0.08
			 & \cellcolor{white!100} 0.77 $\pm$ 0.04 \\

			 & \cellcolor{gray!10} 0.87* $\pm$ 0.04
			 & \cellcolor{gray!10} 0.73 $\pm$ 0.07
			 & \cellcolor{gray!10} 0.68* $\pm$ 0.03
			 & \cellcolor{gray!10} 0.67 $\pm$ 0.03
			 & \cellcolor{gray!10} 0.84* $\pm$ 0.02
			 & \cellcolor{gray!10} 0.83* $\pm$ 0.08
			 & \cellcolor{gray!10} \textbf{0.69} $\pm$ \textbf{0.06}
			 & \cellcolor{gray!10} 0.83* $\pm$ 0.04 \\

\multirow{2}{*}{SYMBA} 
			 & \cellcolor{white!100} \textbf{0.84} $\pm$ \textbf{0.04}
			 & \cellcolor{white!100} \textbf{0.71} $\pm$ \textbf{0.07}
			 & \cellcolor{white!100} \textbf{0.56} $\pm$ \textbf{0.05}
			 & \cellcolor{white!100} 0.62 $\pm$ 0.04
			 & \cellcolor{white!100} \textbf{0.83} $\pm$ \textbf{0.03}
			 & \cellcolor{white!100} \textbf{0.73} $\pm$ \textbf{0.08}
			 & \cellcolor{white!100} 0.48 $\pm$ 0.05
			 & \cellcolor{white!100} \textbf{0.80} $\pm$ \textbf{0.03} \\

			 & \cellcolor{gray!10} \textbf{0.91} $\pm$ \textbf{0.03}
			 & \cellcolor{gray!10} \textbf{0.76} $\pm$ \textbf{0.07}
			 & \cellcolor{gray!10} \textbf{0.72} $\pm$ \textbf{0.03}
			 & \cellcolor{gray!10} \textbf{0.68} $\pm$ \textbf{0.05}
			 & \cellcolor{gray!10} \textbf{0.91} $\pm$ \textbf{0.02}
			 & \cellcolor{gray!10} \textbf{0.88} $\pm$ \textbf{0.07}
			 & \cellcolor{gray!10} 0.68 $\pm$ 0.04
			 & \cellcolor{gray!10} \textbf{0.88} $\pm$ \textbf{0.03} \\

\bottomrule
\end{tabular}
\end{adjustbox}
\end{table*}

We next analyzed the contribution of the individual components of the proposed SYMBA masking strategy through a leave-one-out ablation design, in which each variant removed a single constraint from the final SYMBA masking pipeline while keeping every other component fixed. This isolated the marginal contribution of each constraint independently, rather than assuming a fixed order in which they had to be introduced.

As shown in Table~\ref{tab:masking_ablation}, removing the anatomical shape-prior term (\textit{no shapes}) produced the broadest and most consistent degradation across tasks, and was the only change that reached statistical significance in most cases. The effect was largest on ADNI Sex classification (AUROC dropped from 0.94 to 0.89*, BACC from 0.86 to 0.81*) and ADNI 3-Class classification (AUROC 0.74 to 0.70*), with a smaller but still significant drop on ADNI AD vs CN AUROC (0.94 to 0.90*) and on AIBL Sex AUROC (0.92 to 0.88*). 

Removing the background removal, hemispheric-symmetry, and minimum-tissue-fraction constraints together (anatomical validity constraints off) had a smaller and more selective effect. On ADNI AD vs CN and sMCI vs pMCI, performance was statistically indistinguishable from the full model on both metrics, and on several AIBL tasks (Sex BACC, AD vs CN AUROC, and 3-Class AUROC) the ablated variant performs on par with, or nominally exceeded, full SYMBA (e.g.\ AIBL AD vs CN AUROC of 0.90 vs\ 0.88 for the full model). The only tasks where its removal was clearly detrimental were ADNI 3-Class and Amyloid, where the full model retained a modest edge. 

\begin{table*}[ht!]
\centering
\caption{Ablation of anatomically constrained masking components within SYMBA. Each variant turned off one of the constraints of the final SYMBA configuration pipeline, enabling assessment of their marginal contribution to downstream performance on ADNI and AIBL. Results were reported as mean $\pm$ standard deviation across 10 cross-validation folds. For each variant, the top row (white background) indicates Balanced Accuracy (BACC) and the bottom row (gray background) indicates AUROC (AUC). Bold indicates the best performance per task, and asterisk ($^*$) denotes statistically significant difference ($p < 0.05$) compared to the baseline (SYMBA).}
\label{tab:masking_ablation}
\setlength{\tabcolsep}{2pt}
\begin{adjustbox}{max width=\textwidth}
\begin{tabular}{ccccccccc}
\toprule
& \multicolumn{5}{c}{\textbf{ADNI}} & \multicolumn{3}{c}{\textbf{AIBL}} \\
\cmidrule(lr){2-6}\cmidrule(l){7-9}
\textbf{Constraints} & \textbf{AD vs CN} & \textbf{sMCI vs pMCI} & \textbf{3-Class} & \textbf{Amyloid} & \textbf{Sex} & \textbf{AD vs CN} & \textbf{3-Class} & \textbf{Sex} \\ \midrule
\multirow{2}{*}{---} 
			 & \cellcolor{white!100} \textbf{0.86} $\pm$ \textbf{0.04}
			 & \cellcolor{white!100} 0.66 $\pm$ 0.08
			 & \cellcolor{white!100} 0.55 $\pm$ 0.03
			 & \cellcolor{white!100} 0.65 $\pm$ 0.04
			 & \cellcolor{white!100} \textbf{0.86} $\pm$ \textbf{0.02}
			 & \cellcolor{white!100} 0.76 $\pm$ 0.10
			 & \cellcolor{white!100} 0.49 $\pm$ 0.09
			 & \cellcolor{white!100} \textbf{0.86} $\pm$ \textbf{0.04} \\

			 & \cellcolor{gray!10} \textbf{0.93} $\pm$ \textbf{0.02}
			 & \cellcolor{gray!10} 0.73 $\pm$ 0.07
			 & \cellcolor{gray!10} 0.72 $\pm$ 0.02
			 & \cellcolor{gray!10} 0.69 $\pm$ 0.04
			 & \cellcolor{gray!10} \textbf{0.93} $\pm$ \textbf{0.01}
			 & \cellcolor{gray!10} \textbf{0.90} $\pm$ \textbf{0.08}
			 & \cellcolor{gray!10} 0.71 $\pm$ 0.07
			 & \cellcolor{gray!10} \textbf{0.93} $\pm$ \textbf{0.02} \\
\multirow{2}{*}{Shapes} 
			 & \cellcolor{white!100} \textbf{0.87} $\pm$ \textbf{0.04}
			 & \cellcolor{white!100} \textbf{0.71} $\pm$ \textbf{0.05}
			 & \cellcolor{white!100} 0.57 $\pm$ 0.03
			 & \cellcolor{white!100} 0.63 $\pm$ 0.04
			 & \cellcolor{white!100} 0.86 $\pm$ 0.02
			 & \cellcolor{white!100} 0.77 $\pm$ 0.08
			 & \cellcolor{white!100} 0.49 $\pm$ 0.07
			 & \cellcolor{white!100} \textbf{0.84} $\pm$ \textbf{0.04} \\
			 & \cellcolor{gray!10} \textbf{0.94} $\pm$ \textbf{0.03}
			 & \cellcolor{gray!10} \textbf{0.78} $\pm$ \textbf{0.06}
			 & \cellcolor{gray!10} 0.73 $\pm$ 0.03
			 & \cellcolor{gray!10} 0.68 $\pm$ 0.03
			 & \cellcolor{gray!10} 0.93 $\pm$ 0.01
			 & \cellcolor{gray!10} \textbf{0.90} $\pm$ \textbf{0.08}
			 & \cellcolor{gray!10} \textbf{0.71} $\pm$ \textbf{0.06}
			 & \cellcolor{gray!10} \textbf{0.92} $\pm$ \textbf{0.02} \\
\multirow{2}{*}{Anatomical} 
			 & \cellcolor{white!100} 0.83 $\pm$ 0.04
			 & \cellcolor{white!100} 0.69 $\pm$ 0.08
			 & \cellcolor{white!100} 0.52 $\pm$ 0.04
			 & \cellcolor{white!100} \textbf{0.65} $\pm$ \textbf{0.04}
			 & \cellcolor{white!100} 0.81* $\pm$ 0.02
			 & \cellcolor{white!100} 0.77 $\pm$ 0.10
			 & \cellcolor{white!100} 0.46 $\pm$ 0.09
			 & \cellcolor{white!100} 0.80 $\pm$ 0.06 \\
			 & \cellcolor{gray!10} 0.90* $\pm$ 0.03
			 & \cellcolor{gray!10} 0.76 $\pm$ 0.08
			 & \cellcolor{gray!10} 0.70* $\pm$ 0.03
			 & \cellcolor{gray!10} 0.68 $\pm$ 0.04
			 & \cellcolor{gray!10} 0.89* $\pm$ 0.02
			 & \cellcolor{gray!10} 0.86 $\pm$ 0.07
			 & \cellcolor{gray!10} 0.68 $\pm$ 0.04
			 & \cellcolor{gray!10} 0.88* $\pm$ 0.04 \\
\multirow{2}{*}{Shapes+Anatomical} 
			 & \cellcolor{white!100} \textbf{0.87} $\pm$ \textbf{0.04}
			 & \cellcolor{white!100} \textbf{0.71} $\pm$ \textbf{0.06}
			 & \cellcolor{white!100} \textbf{0.59} $\pm$ \textbf{0.03}
			 & \cellcolor{white!100} \textbf{0.65} $\pm$ \textbf{0.03}
			 & \cellcolor{white!100} \textbf{0.86} $\pm$ \textbf{0.02}
			 & \cellcolor{white!100} \textbf{0.78} $\pm$ \textbf{0.08}
			 & \cellcolor{white!100} \textbf{0.51} $\pm$ \textbf{0.09}
			 & \cellcolor{white!100} 0.83 $\pm$ 0.03 \\
			 & \cellcolor{gray!10} \textbf{0.94} $\pm$ \textbf{0.03}
			 & \cellcolor{gray!10} \textbf{0.78} $\pm$ \textbf{0.08}
			 & \cellcolor{gray!10} \textbf{0.74} $\pm$ \textbf{0.02}
			 & \cellcolor{gray!10} \textbf{0.70} $\pm$ \textbf{0.04} 
			 & \cellcolor{gray!10} \textbf{0.94} $\pm$ \textbf{0.01}
			 & \cellcolor{gray!10} 0.88 $\pm$ 0.09
			 & \cellcolor{gray!10} \textbf{0.71} $\pm$ \textbf{0.08}
			 & \cellcolor{gray!10} \textbf{0.92} $\pm$ \textbf{0.03} \\
\bottomrule
\end{tabular}
\end{adjustbox}
\end{table*}

\subsubsection{Reconstruction objective}

\begin{table*}[ht!]
\centering
\caption{Effect of reconstruction objective on downstream performance within the SYMBA framework, comparing voxel-wise losses (MSE), structural similarity-based objectives (SSIM), hybrid formulations, and MS-SSIM supervision. Results are reported as mean $\pm$ standard deviation across 10 cross-validation folds. For each variant, the top row (white background) indicates Balanced Accuracy (BACC) and the bottom row (gray background) indicates AUROC (AUC). Bold indicates the best performance per task, and asterisk ($^*$) denotes a statistically significant difference ($p < 0.05$).}
\label{tab:SSLobjective}
\setlength{\tabcolsep}{2pt}
\begin{adjustbox}{max width=\textwidth}
\begin{tabular}{ccccccccc}
\toprule
& \multicolumn{5}{c}{\textbf{ADNI}} & \multicolumn{3}{c}{\textbf{AIBL}} \\
\cmidrule(lr){2-6}\cmidrule(l){7-9}
\textbf{Loss} & \textbf{AD vs CN} & \textbf{sMCI vs pMCI} & \textbf{3-Class} & \textbf{Amyloid} & \textbf{Sex} & \textbf{AD vs CN} & \textbf{3-Class} & \textbf{Sex} \\ \midrule

\multirow{2}{*}{MSE} 
			 & \cellcolor{white!100} 0.86 $\pm$ 0.04
			 & \cellcolor{white!100} 0.70 $\pm$ 0.06
			 & \cellcolor{white!100} 0.56* $\pm$ 0.05
			& \cellcolor{white!100} 0.64 $\pm$ 0.04
			 & \cellcolor{white!100} 0.86 $\pm$ 0.02
			 & \cellcolor{white!100} 0.77 $\pm$ 0.08
			 & \cellcolor{white!100} 0.49 $\pm$ 0.06
			 & \cellcolor{white!100} 0.81 $\pm$ 0.04 \\

			 & \cellcolor{gray!10} 0.93* $\pm$ 0.02
			 & \cellcolor{gray!10} \textbf{0.78} $\pm$ \textbf{0.06}
			 & \cellcolor{gray!10} 0.72* $\pm$ 0.03
			 & \cellcolor{gray!10} \textbf{0.71} $\pm$ \textbf{0.04}
			 & \cellcolor{gray!10} 0.93 $\pm$ 0.02
			 & \cellcolor{gray!10} 0.88 $\pm$ 0.07
			 & \cellcolor{gray!10} 0.70 $\pm$ 0.06
			 & \cellcolor{gray!10} 0.89* $\pm$ 0.04 \\
\multirow{2}{*}{SSIM} 
			 & \cellcolor{white!100} 0.85 $\pm$ 0.04
			 & \cellcolor{white!100} 0.70 $\pm$ 0.05
			 & \cellcolor{white!100} 0.56 $\pm$ 0.03
			 & \cellcolor{white!100} 0.63 $\pm$ 0.04
			 & \cellcolor{white!100} 0.85 $\pm$ 0.02
			 & \cellcolor{white!100} 0.74 $\pm$ 0.06
			 & \cellcolor{white!100} 0.50 $\pm$ 0.07
			 & \cellcolor{white!100} 0.81 $\pm$ 0.05 \\

			 & \cellcolor{gray!10} 0.93* $\pm$ 0.03
			 & \cellcolor{gray!10} 0.74* $\pm$ 0.06
			 & \cellcolor{gray!10} 0.73* $\pm$ 0.02
			 & \cellcolor{gray!10} 0.69 $\pm$ 0.03
			 & \cellcolor{gray!10} 0.93 $\pm$ 0.01
			 & \cellcolor{gray!10} \textbf{0.90} $\pm$ \textbf{0.07}
			 & \cellcolor{gray!10} \textbf{0.75} $\pm$ \textbf{0.05}
			 & \cellcolor{gray!10} 0.89* $\pm$ 0.04 \\

\multirow{2}{*}{SSIM l1} 
			 & \cellcolor{white!100} 0.86 $\pm$ 0.04
			 & \cellcolor{white!100} 0.69 $\pm$ 0.07
			 & \cellcolor{white!100} 0.56* $\pm$ 0.04
			 & \cellcolor{white!100} 0.63 $\pm$ 0.04
			 & \cellcolor{white!100} 0.86 $\pm$ 0.02
			 & \cellcolor{white!100} 0.77 $\pm$ 0.09
			 & \cellcolor{white!100} 0.48 $\pm$ 0.07
			 & \cellcolor{white!100} \textbf{0.83} $\pm$ \textbf{0.04} \\

			 & \cellcolor{gray!10} 0.92* $\pm$ 0.02
			 & \cellcolor{gray!10} 0.76 $\pm$ 0.06
			 & \cellcolor{gray!10} 0.72* $\pm$ 0.03
			 & \cellcolor{gray!10} 0.69 $\pm$ 0.03
			 & \cellcolor{gray!10} 0.93 $\pm$ 0.01
			 & \cellcolor{gray!10} 0.89 $\pm$ 0.07
			 & \cellcolor{gray!10} 0.70 $\pm$ 0.04
			 & \cellcolor{gray!10} 0.90 $\pm$ 0.03 \\

\multirow{2}{*}{MS-SSIM} 
			 & \cellcolor{white!100} \textbf{0.87} $\pm$ \textbf{0.04}
			 & \cellcolor{white!100} \textbf{0.71} $\pm$ \textbf{0.06}
			 & \cellcolor{white!100} \textbf{0.59} $\pm$ \textbf{0.03}
			 & \cellcolor{white!100} \textbf{0.65} $\pm$ \textbf{0.03}
			 & \cellcolor{white!100} \textbf{0.86} $\pm$ \textbf{0.02}
			 & \cellcolor{white!100} \textbf{0.78} $\pm$ \textbf{0.10}
			 & \cellcolor{white!100} \textbf{0.51} $\pm$ \textbf{0.09}
			 & \cellcolor{white!100} 0.83 $\pm$ 0.03 \\

			& \cellcolor{gray!10} \textbf{0.94} $\pm$ \textbf{0.03}
			 & \cellcolor{gray!10} 0.78 $\pm$ 0.08
			 & \cellcolor{gray!10} \textbf{0.74} $\pm$ \textbf{0.02}
			 & \cellcolor{gray!10} 0.70 $\pm$ 0.04
			 & \cellcolor{gray!10} \textbf{0.94} $\pm$ \textbf{0.01}
			 & \cellcolor{gray!10} 0.88 $\pm$ 0.09
			 & \cellcolor{gray!10} 0.71 $\pm$ 0.08
			 & \cellcolor{gray!10} \textbf{0.92} $\pm$ \textbf{0.03} \\

\bottomrule
\end{tabular}
\end{adjustbox}
\end{table*}

We investigated the influence of the reconstruction objective on the quality of the learned representations by comparing four commonly used losses within the proposed framework: voxel-wise Mean Squared Error (MSE), Structural Similarity (SSIM), a hybrid SSIM+L1 objective, and the MS-SSIM.

Table~\ref{tab:SSLobjective} showed that the reconstruction objective had a substantial impact on representation transferability. No single loss achieved the highest performance on every downstream task, indicating that each objective promoted different inductive biases during pretraining. MSE performed competitively on several ADNI tasks, achieving the highest performance AUC for MCI progression prediction and amyloid status.

In contrast, objectives based on structural similarity exhibited better generalization across datasets. %While SSIM alone produced results comparable to MSE on ADNI, the 
MS-SSIM objective achieved the highest Balanced accuracy across all ADNI tasks, including 3-Class disease staging and AD vs CN classification, while also securing top AUC scores for AD vs CN, 3-Class diagnosis and Sex prediction. On the independent AIBL cohort, MS-SSIM led in Balanced Accuracy for AD vs CN and 3-Class diagnosis, as well as Sex prediction. However, standard single-scale SSIM achieved the highest AUC for AIBL AD vs CN and 3-Class diagnosis. 

Interestingly, simply combining structural and voxel-wise supervision through the SSIM+L1 objective did not produce consistent improvements over either individual loss, with the exception of top Balanced Accuracy in AIBL Sex prediction. This indicated that the benefits of structural reconstruction were not obtained by adding a pixel-wise penalty, but rather by explicitly encouraging the model to preserve anatomical organization during reconstruction.

\subsection{Reconstruction analysis}

\begin{figure}[H]
    \centering
    \includegraphics[width=0.99\linewidth]%{Figures/brains_masked_adni.png} 
    {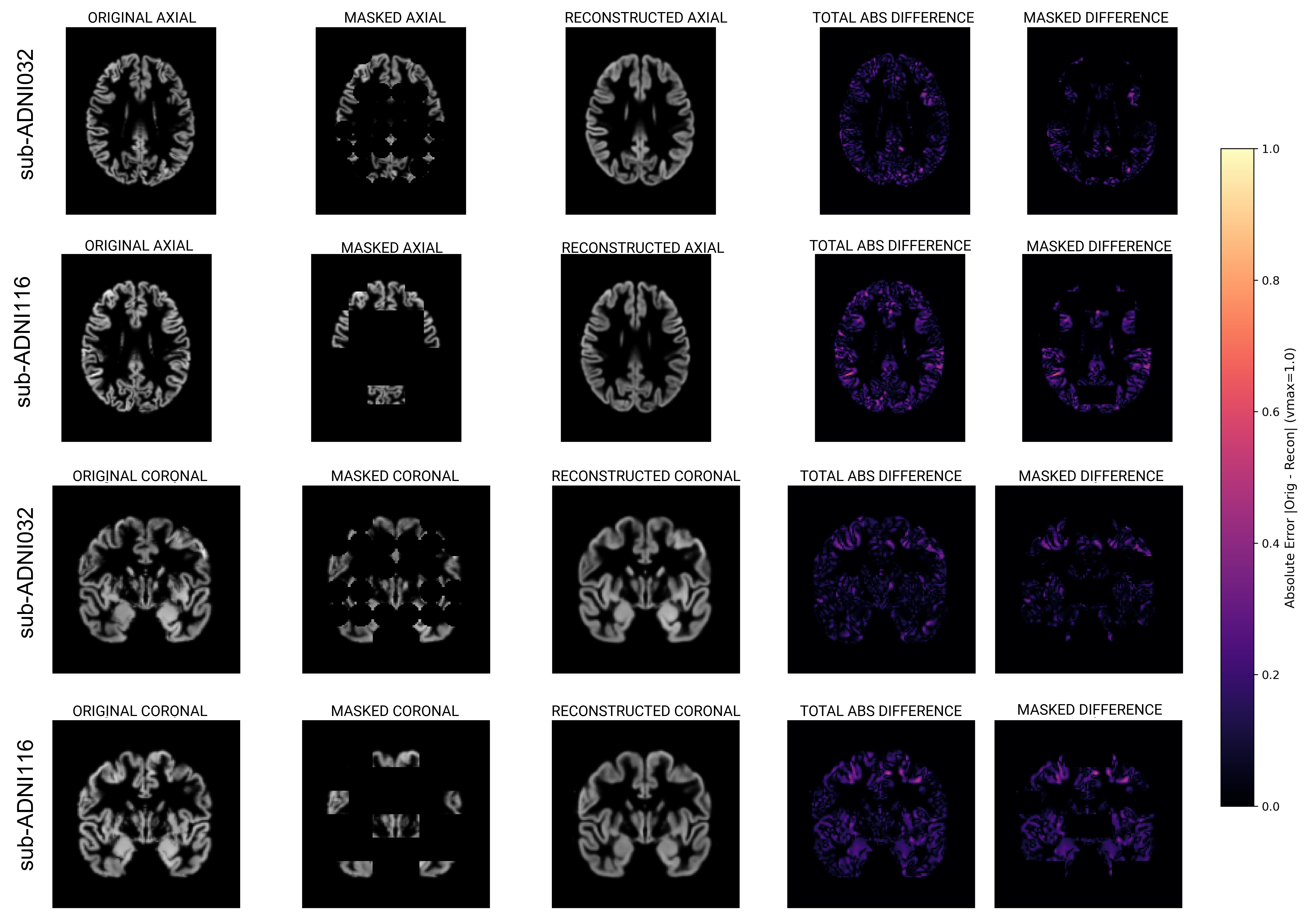} 
    %Figures/brains_spheres.png
    \caption{Representative reconstructions produced ResNet under the masking ratio (75\%) and patch size 16 for the axial and coronal view. Columns correspond to the original image, masked input, reconstructed image, total absolute difference and masked difference respectively.}
    \label{fig:brains}
\end{figure}

To qualitatively assess the effect of the encoder architecture and parameters on the reconstruction process, Figure~\ref{fig:brains} shows two representative reconstructions obtained using the final SYMBA model under masking ratio of 75\% and patch size 16. The example subjects were selected at random from the held-out test set without prior visual inspection to ensure unbiased sampling. While the quantitative evaluation throughout this work focused on the transferability of the learned representations rather than reconstruction quality itself, visual inspection provided additional insight into the information encoded in the latent space. 
Specifically, by comparing the original volume, the 75\% masked input, and the final reconstruction, we observed that the network successfully captured global contextual dependencies. Despite the heavily fragmented input, the model accurately recovered the overall anatomical structure and major boundary continuities.

\section{Discussion}

This study investigated whether incorporating anatomical and morphological priors into self-supervised pretraining can improve the transferability of structural brain MRI representations for neurodegenerative disease analysis. The results suggest that the design of the pretraining objective plays an important role in determining the information retained in the learned representations. In particular, the performance of SYMBA across independent cohorts, heterogeneous downstream tasks, and limited-label settings supports the value of reconstruction objectives that encourage the modeling of spatial relationships and global brain morphology. Beyond the overall improvements in predictive performance, our findings provide insights into the relationship between self-supervised objectives, anatomical inductive biases, and downstream representation quality.

\subsection{Learning Morphological Representations for Neurodegenerative Disease Analysis}

An important question in self-supervised neuroimaging is which anatomical properties a representation should preserve to support subsequent clinical analysis. Unlike tasks such as image registration or segmentation, which emphasize accurate spatial correspondence or local boundary delineation, neurodegenerative disease classification requires identifying structural variations that may be subtle, distributed, and spatially interdependent. This distinction has implications for the design of self-supervised objectives, since representations optimized for one anatomical task may not necessarily retain the information most relevant to another.

The differences observed between SYMBA and the evaluated pretrained models are consistent with this interpretation. In contrast to objectives emphasizing instance discrimination, anatomical correspondence, or segmentation, SYMBA requires reconstructing missing tissue from the surrounding anatomical context. This encourages the encoder to retain information that supports the recovery of spatial relationships between brain structures, rather than optimizing exclusively for local features or similarity between augmented views. The progressive masking curriculum further introduces reconstruction problems at different spatial scales, potentially promoting representations that capture both localized anatomical variations and larger morphological patterns.

The differences across downstream tasks also provide insight into the information captured by these representations. While SYMBA performed well on established diagnostic categories, improvements were less pronounced for MCI progression and amyloid status prediction. This distinction is relevant because sMRI provides an indirect characterization of the underlying pathological processes. Clinical progression and molecular pathology cannot necessarily be inferred from macroscopic morphology with the same reliability as established disease-related anatomical differences. Consequently, the benefits of improved structural representation learning may be constrained by the information intrinsically available in the imaging modality.

\subsection{Transferability and Label Efficiency}

A central motivation for SSL in neuroimaging is the possibility of learning representations from heterogeneous unlabeled populations and transferring them to clinical cohorts where annotated data are limited. SYMBA was pretrained without access to either ADNI or AIBL, yet its representations supported multiple downstream classification tasks using frozen encoders and linear classifiers. This suggests that the pretrained latent space contains anatomical information that remains useful across populations and acquisition settings without requiring extensive task-specific adaptation.

The use of frozen linear probing is particularly relevant to this interpretation. The evaluation reduces the extent to which differences in performance can be attributed to task-specific representation adaptation due to the restriction of downstream learning to a linear classifier. The resulting improvements therefore support the usefulness of the anatomical information encoded during pretraining. Nevertheless, this protocol evaluates linear accessibility of downstream information rather than the full potential of each architecture after fine-tuning.

The low-data experiments provide further evidence of this property. SYMBA's advantage was most pronounced when very few labeled examples were available, suggesting that much of the information needed for downstream discrimination was already accessible in the pretrained feature space. From a practical perspective, this is particularly relevant to neuroimaging applications where obtaining reliable labels requires expensive biomarker measurements, specialist assessments, or prolonged clinical follow-up.

However, the different data efficiency curves reveal that the value of self-supervised pretraining is not uniform across tasks. For several ADNI classification problems, supervised models progressively approached the performance of SYMBA as more labeled examples became available. For MCI progression prediction, the supervised baseline eventually surpassed the frozen representations. One possible explanation is that pretraining captures broadly useful anatomical characteristics, whereas distinguishing stable from progressive MCI may benefit from learning more specific combinations of structural features that are not linearly accessible in the pretrained space.

These observations suggest a complementary relationship between general-purpose pretraining and task-specific optimization. Frozen representations may be particularly attractive in severely data-constrained settings, whereas fine-tuning could become beneficial when larger annotated cohorts are available or when the prediction target requires more specialized information. Investigating this trade-off would help determine when the additional complexity of end-to-end adaptation is justified.

\subsection{Anatomical Inductive Biases and Representation Quality}

The ablation experiments offer additional insight into how architectural and methodological decisions influence representation learning. An important observation is that improvements do not arise uniformly from increasing architectural complexity or imposing additional anatomical constraints. Instead, the findings suggest a trade-off between preserving detailed spatial information and encouraging the encoder to learn compact representations of global brain morphology.

This trade-off is illustrated by the comparison between residual autoencoders and U-Net architectures. Although skip connections can facilitate reconstruction by transferring spatial information directly between encoder and decoder, they also provide pathways that bypass the latent bottleneck. In a representation-learning setting, this may reduce the need for the encoder to retain all the information required for reconstruction. The stronger performance of residual architectures on several disease-related tasks is consistent with this hypothesis, although the competitive performance of U-Net on other endpoints indicates that preserving fine-grained spatial detail can also be beneficial. Importantly, these findings do not establish that skip connections are universally detrimental, but rather suggest that reconstruction fidelity and downstream representation quality are not necessarily aligned.

The masking ablations reinforce the importance of the reconstruction problem presented to the encoder. The benefits associated with geometrically structured masking suggest that the spatial organization of missing information influences which anatomical dependencies the network learns. However, the less consistent contribution of additional masking constraints indicates that anatomical priors should not be assumed to be beneficial simply because they encode biological knowledge. Their usefulness depends on whether they encourage the model to preserve information relevant to the downstream task without unnecessarily restricting the variability encountered during pretraining.

A similar interpretation applies to the reconstruction objective. Whereas voxel-wise losses prioritize local intensity agreement, structural similarity objectives evaluate relationships involving local contrast and spatial organization. The overall performance of MS-SSIM is consistent with the hypothesis that multiscale structural reconstruction provides useful supervision for morphometric representation learning. Nevertheless, the competitive results obtained with MSE and single-scale SSIM on certain tasks indicate that different reconstruction criteria may favor different aspects of anatomical information. This raises the possibility that reconstruction objectives could be adapted to the characteristics of the intended downstream application rather than selected solely according to reconstruction accuracy.

Taken together, these findings suggest that effective self-supervised learning for structural neuroimaging requires balancing anatomical detail, spatial context, and latent compression. Rather than maximizing architectural capacity or reconstruction fidelity independently, the objective should be to encourage representations that retain informative morphological variation while remaining transferable across datasets and tasks.

\subsection{Limitations and Future Directions}

Several limitations should be considered when interpreting these findings. First, although the evaluation included two independent cohorts excluded from SYMBA pretraining, the experiments do not explicitly isolate the effects of scanner characteristics, acquisition protocols, or population differences. Furthermore, some competing models were pretrained using data from the evaluation cohorts, complicating direct comparisons of their exposure to the target distributions. Additional experiments involving site-stratified evaluation or controlled acquisition shifts would be necessary to characterize robustness to these factors more precisely.

Second, downstream evaluation relied primarily on frozen linear probing. While this provides a controlled means of assessing representation quality, it does not establish the maximum predictive performance achievable through adaptation. The results for MCI progression prediction particularly motivate investigating whether partial fine-tuning or parameter-efficient adaptation can complement the pretrained representations.

Third, the present framework is restricted to structural gray matter maps derived from T1-weighted MRI. Although these images provide valuable information about neurodegeneration, they cannot capture all aspects of disease pathology. Integrating complementary imaging modalities or clinical information could help address prediction problems in which morphological information alone is insufficient, particularly molecular biomarker estimation and early disease progression.

Finally, the analysis uses baseline structural images, without explicitly modeling within-subject changes over time. Although the proposed framework learns representations associated with subsequent clinical progression, it does not directly characterize longitudinal neurodegenerative trajectories. Extending anatomical self-supervision to repeated observations could provide a more direct means of learning how structural changes evolve, provided suitable longitudinal data are available.

Overall, this work supports anatomically informed reconstruction as a promising strategy for self-supervised structural MRI representation learning. More broadly, the findings emphasize that pretraining objectives should be designed around the anatomical information relevant to the intended downstream applications, while recognizing the trade-offs between general-purpose transferability and task-specific adaptation.

\section{Conclusions}
\label{sec:conclusions}
In this work, we presented SYMBA, a self-supervised pretraining framework systematically optimized for structural brain MRI analysis, demonstrating how neuroanatomical inductive biases yield highly generalizable representations tailored for neurodegenerative disease analysis. Through a rigorous evaluation pipeline, we demonstrate that anatomically-informed generative pretraining produces more transferable representations than existing SSL approaches for sMRI. Rather than adapting objectives originally developed for natural images, SYMBA incorporates neuroanatomical priors into the masking strategy, reconstruction objective, and training curriculum, encouraging the encoder to model structural brain organization across multiple spatial scales.

%To the best of our knowledge, this work is the first to strictly evaluate FM performance via linear probing rather than full downstream fine-tuning. 
The deliberate methodological choice of using linear probing instead of fine-tuning allowed for a rigorous evaluation of the intrinsic quality of the learned latent space, removing the confounding variables and artifacts typically introduced by complex fine-tuning schedulers. Despite this conservative evaluation protocol, SYMBA demonstrated superior performance and out-of-domain generalizability across a diverse array of clinical tasks when evaluated on two independent and unseen external datasets, ADNI and AIBL.

The clinical applicability of this approach is more notable in our low-data regime evaluations. In extremely low-data settings, SYMBA achieved an AUROC close to 0.90 for Alzheimer's disease classification using only six training samples (1\% of the training data), drastically outperforming fully supervised baselines trained from scratch. This result indicates that much of the disease-discriminative information is already encoded during pretraining, substantially reducing the dependence on large annotated cohorts.
% and overall matching or exceeding massive FMs that have required substantially larger pretraining resources. By prioritizing a phased, step-by-step optimization of embedding representations prior to downstream adaptation, this methodology has addressed a critical bottleneck in medical AI. These findings aim to establish a highly data-efficient foundation capable of generalizing to unseen datasets and rare neurological conditions in which labeled training samples are highly scarce. 
Our findings also suggest that for structural MRI the explicit adaptation of the SSL objective to the spatial organization of brain morphology is critical. 
Future work will investigate parameter-efficient adaptation strategies, multimodal downstream tasks, and transfer to rarer neurological disorders where labeled data remain extremely limited.
% Future work will extend this baseline to explore specialized fine-tuning paradigms to obtain the full potential of these robust structural embeddings and observe the accuracy in additional downstream tasks for rare diseases from scarce available labeled data.

\section*{Acknowledgments}
Data collection and sharing for this project was funded by the Alzheimer’s Disease Neuroimaging Initiative (ADNI) (National Institutes of Health Grant U19AG024904). The grantee organization is the Northern California Institute for Research and Education, United States.ADNI is funded by the National Institute on Aging, United States, theNational Institute of Biomedical Imaging and Bioengineering, UnitedStates, and through generous contributions from the following: AbbVie, Alzheimer’s Association; Alzheimer’s Drug Discovery Foundation; Ara-clon Biotech; BioClinica, Inc.; Biogen; Bristol-Myers Squibb Company; CereSpir, Inc.; Cogstate; Eisai Inc.; Elan Pharmaceuticals, Inc.; Eli Lillyand Company; EuroImmun; F. Hoffmann-La Roche Ltd and its affiliatedcompany Genentech, Inc.; Fujirebio; GE Healthcare; IXICO Ltd.; JanssenAlzheimer Immunotherapy Research \& Development, LLC.; Johnson\& Johnson Pharmaceutical Research \& Development LLC.; Lumosity; Lundbeck; Merck \& Co., Inc.; Meso Scale Diagnostics, LLC.; NeuroRxResearch; Neurotrack Technologies; Novartis Pharmaceuticals Corpo-ration; Pfizer Inc.; Piramal Imaging; Servier; Takeda PharmaceuticalCompany; and Transition Therapeutics. The Canadian Institutes of Health Research is providing funds to support ADNI clinical sites in Canada. Private sector contributions are facilitated by the Foundation for the National Institutes of Health (www.fnih.org). The study is coordinated by the Alzheimer’s Therapeutic Research Institute at the University of Southern California, and data are disseminated by the Laboratory for Neuro Imaging at the University of Southern California.
Data were provided in part by OASIS [Longitudinal Multimodal Neuroimaging: Principal Investigators: T. Benzinger, D. Marcus, J. Morris; NIH P30 AG066444, P50 AG00561, P30 NS09857781, P01 AG026276, P01 AG003991, R01 AG043434, UL1 TR000448, R01 EB009352. AV-45 doses were provided by Avid Radiopharmaceuticals, a wholly owned subsidiary of Eli Lilly]
IXI dataset data collection was supported by the Engineering and Physical Sciences Research Council (EPSRC) under grant GR/S21533/02.
Part of the Data collection and sharing for this project (CamCAN database) was provided by the Cambridge Centre for Ageing and Neuroscience. Funding for CamCAN was provided by the UK Biotechnology and Biological Sciences Research Council (grant number BB/H008217/1), along with support from the UK Medical Research Council and University of Cambridge, UK.

CSS and VGV acknowledge financial support from grant TEC-2024/COM-89 funded by the Autonomous Community of Madrid. CSS is also supported by the Comunidad de Madrid through the 2025 César Nombela programme (Grant 2025-T1/COM-36091) and by grant PID2025-174087OA-I00 funded by MCIN/AEI/10.13039/501100011033 and ERDF/EU. VGV is partially supported by grant PID2023-146684NB-I00 funded by MCIN/AEI/10.13039/501100011033 and ERDF/EU. VGV is an ELLIS Unit Madrid member funded by the Community of Madrid. ABL acknowledges financial support from the Helmut Horten Foundation (Clinical Translational Grant 2024-CON-024).JT and EM acknowledge support by the Research Council of Finland under grants 346934 (PRIMAL), and  358944 (Flagship of Advanced Mathematics for Sensing Imaging and Modeling). NBE acknowledges financial support from "la Caixa" Foundation (ID 100010434) with fellowship code LCF/BQ/DFR26/12001117. MRI processing was performed on servers provided by UEF Bioinformatics Center, Biocenter Kuopio, Biocenter Finland, University of Eastern Finland, Finland. 

\footnotesize
\bibliography{References_diferent_versions/refs_final}
\bibliographystyle{cas-model2-names}

% \clearpage
\appendix

\end{document}